\documentclass[10pt,twocolumn,letterpaper]{article}

\usepackage[pagenumbers]{wacv} 

\newcommand{\errormap}{E}

\newcommand{\edgemap}{M^{\text{edg}}}
\newcommand{\objectmask}{M^{\text{sub}}}
\newcommand{\totalobjects}{N}
\newcommand{\totalframes}{F}
\newcommand{\oid}{n}
\newcommand{\pointtrack}{T}
\newcommand{\pid}{p}
\newcommand{\totalpoints}{P}
\newcommand{\trackdist}{d}
\newcommand{\diffdist}{\delta}
\newcommand{\norm}[1]{\left\lVert#1\right\rVert}

\usepackage{amsmath}
\usepackage{graphicx} 
\usepackage{caption}
\usepackage{subcaption}
\usepackage{adjustbox}
\usepackage{booktabs} 
\usepackage{nicefrac}
\usepackage{wrapfig}
\usepackage{soul}
\usepackage{enumitem}
\usepackage{colortbl} 
\usepackage{multirow}

\definecolor{wacvblue}{rgb}{0.21,0.49,0.74}
\usepackage[pagebackref,breaklinks,colorlinks,allcolors=wacvblue]{hyperref}
\usepackage[accsupp]{axessibility}

\def\wacvPaperID{593} 
\def\confName{WACV}
\def\confYear{2027}

\title{A Closer Look at Evaluation of Text-to-Video Models Under Camera Motion}

\author{
Nithin C. Babu$^{1*}$ \quad
Aniruddha Mahapatra$^{2}$ \quad
Harsh Rangwani$^{3}$ \\
Rajiv Soundararajan$^{1}$ \quad
Kuldeep Kulkarni$^{3}$\\[0.2cm]
$^{1}$Indian Institute of Science, Bangalore, India \\
$^{2}$Adobe Research, San Jose, USA\\
$^{3}$Adobe Research, Bangalore, India\\[0.1cm]
{\tt\small \{nithinc, rajivs\}@iisc.ac.in, \{anmahapa, hrangwani, kulkulka\}@adobe.com}
}

\begin{document}
\maketitle
{\let\thefootnote\relax\footnotetext{$^{*}$Work done in part while the author was an intern at Adobe Research.

Project Page: \href{https://nithincbabu7.github.io/DynamicEval}{\tt nithincbabu7.github.io/DynamicEval}}}

\begin{figure*}
  \centering
  \includegraphics[width=0.99\linewidth]{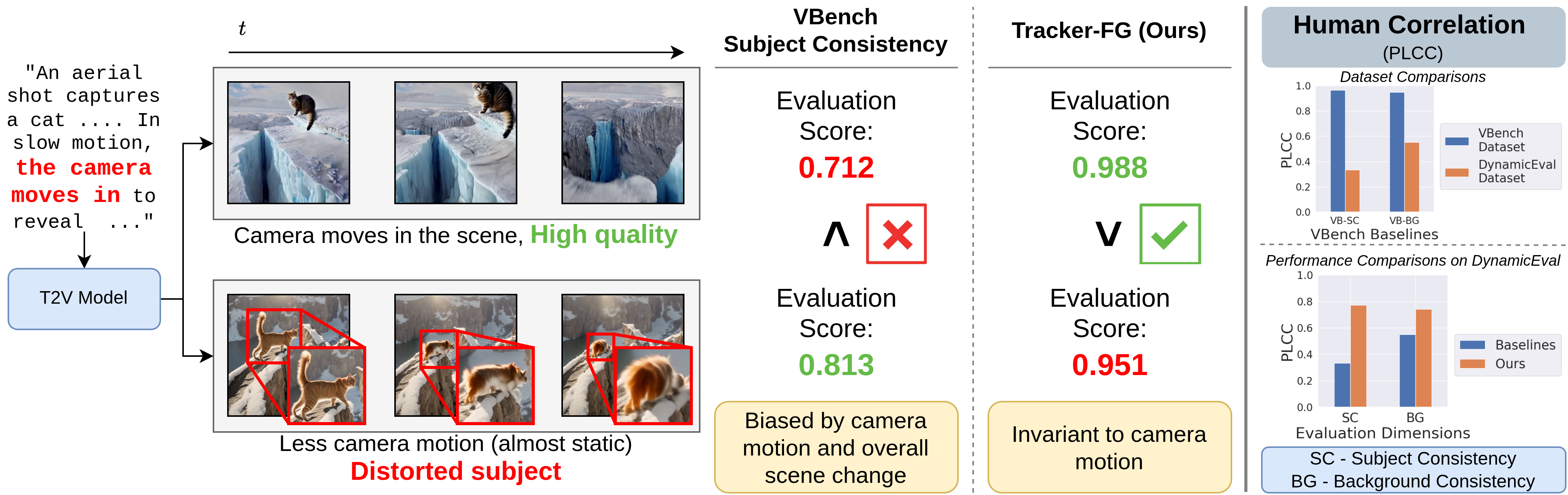}
  \caption{Existing metrics (e.g., VB-Subject Consistency) favor low-motion videos even when they contain subject distortions. Our motion-invariant Tracker-FG metric reliably evaluates foreground quality regardless of camera or object motion. On our large-scale DynamicEval benchmark, prior metrics show substantial performance drops (top right), while our proposed metrics (Tracker-FG and MS-Debias) achieve significantly higher Pearson's Linear Correlation Coefficient (PLCC) with model-level human preferences (bottom right).}
  \label{fig:teaser}
\vspace{-5mm}
\end{figure*}

\begin{abstract}

Existing text-to-video (T2V) evaluation benchmarks, such as VBench and EvalCrafter, predominantly contain either subject-centric prompts or static camera scenes, overlooking the importance of camera motion, essential for producing cinematic shots with realistic scenes. Due to this, the metrics developed in these works often produce incorrect results in the presence of camera motion. To fill this gap, we introduce DynamicEval, a benchmark comprising of systematically curated prompts that emphasize dynamic camera motion, paired with 45k human annotations on video pairs. DynamicEval evaluates two key dimensions of video quality: the consistency of the foreground subjects and the consistency of the background scene. To address the shortcomings of existing metrics under camera motion, we introduce two novel metrics: 1) Tracker-FG, which evaluates subject consistency by assessing the keypoint tracks of the subject. 2) Debiased Motion Smoothness, for background consistency, devoid from issues of occlusions/disocclusions arising from camera and foreground subject movements. Extensive experiments on the large-scale DynamicEval benchmark demonstrate that our proposed metrics achieve stronger correlations with human preferences at both the video level, and the model level with an increase of 0.32 in average Pearson Linear Correlation.

\end{abstract}

\section{Introduction}
\label{sec:intro}
The rapid advancement of foundational text-to-video (T2V) models~\citep{opensora, yang2024cogvideox, kong2024hunyuanvideo, wan2025wan} has necessitated the development of automatic evaluation metrics that correlate highly with human preferences.  
Recent works, such as VBench~\citep{vbench} and EvalCrafter~\citep{evalcrafter}, have introduced evaluation prompt suites along with automatic metrics that assess several dimensions of video quality, including subject consistency, background consistency, text alignment, and color consistency. These benchmarks are designed to provide broad coverage; with their prompts being predominantly generic and subject-centric, and hence overlook the evaluation of text-to-video generation that involves any kind of camera motion. As camera motion gives videos a cinematic, coherent, and realistic feel, there have been attempts in the literature to introduce techniques to generate videos via T2V models with accurate camera motion~\cite{zhang2025recapture, he2024cameractrl, geng2025motion}. 
However, due to lack of prompts having camera motion in the existing benchmarks for T2V models, the automated metrics introduced with them are often biased towards incorrectly rating generated videos with camera motion as low-quality compared to static videos with serious artifacts (Fig.~\ref{fig:teaser}).


To determine the extent of this problem for dynamic scenes in text-to-video (T2V) generation, we introduce DynamicEval, a benchmark designed to complement existing general benchmarks for assessing quality under dynamic camera motion. DynamicEval consists of two key components: (1) a procedurally generated prompt suite featuring highly detailed descriptions that explicitly specify camera motion, and (2) 45k high-quality human annotations across 3k videos generated by ten T2V models using this prompt suite. We conduct a large-scale subjective study in which human annotators compared pairs of videos generated from the same prompt, evaluating them across multiple quality dimensions. We then evaluate existing metrics of VBench~\cite{vbench}, which are most critical for evaluating dynamic videos, namely: 1) VG-SC for foreground (FG) subject consistency and 2) VB-BG for background (BG) consistency. These metrics, which are shown to correlate well with human judgment~\cite{vbench}, drop drastically ($\approx$ 50\%) on the introduced DynamicEval benchmark (Fig.~\ref{fig:teaser}), highlighting their critical design flaws. Further, it is hard to reason about why these metrics do not work as they are either based on global CLIP~\cite{radford2021learning} feature similarities (VB-BG) or DINO features (VB-SC).
To overcome these challenges, we propose fine-grained metrics that are grounded at the pixel level, providing enhanced interpretability for their results.

The other dimensions of video quality, such as temporal flickering, aesthetic quality, and color consistency can still be evaluated similarly with or without camera motion.
Hence for our evaluation, we specifically focus on two critical dimensions that primarily are affected by camera motion: (1) foreground (FG) subject consistency and (2) background (BG) consistency in the generated videos. 

For foreground (FG) subject consistency, the ideal metric should isolate subjects, remain robust to camera/subject motion, and capture long-term subject details. 
The VBench subject consistency metric, which leverages DINO feature similarity across frames, fails to capture these nuances (Fig.~\ref{fig:teaser}).
We propose a fundamentally different approach to measure FG subject consistency by tracking multiple points on the foreground subjects using CoTracker~\citep{cotracker} and monitoring their nearest neighbors. Our subject consistency metric is then defined by analyzing the smoothness of distances between these tracked points over time (Fig.~\ref{fig:FG_SAM_Cotracker}).
This method is highly effective in capturing subtle deformations of the subject throughout the video, which ensures long-term temporal understanding. 

For background (BG) consistency, we first investigate the pitfalls in common evaluation metrics, in particular, VBench Motion Smoothness (VB-MS)~\citep{vbench}. 
Motion Smoothness is based on the accuracy of reconstruction of the mid frame, given the three adjacent frames via optical flow, and hence also provides a pixel-level reconstruction map for evaluation.
However, we observe that this metric accounts for the entire frame including foreground subjects, and yields large errors near occlusions and disocclusions caused by camera motion (Fig.~\ref{fig:MS_debias}).
We overcome these limitations by introducing debiased Motion Smoothness (MS-Debias) , which isolates the effect of foreground objects while removing occlusion-related background pixels. This ensures that our background consistency metric is  temporally stable measure, that is independent of moving subjects and errors arising from camera motion. 
Using our DynamicEval benchmark, we demonstrate that our proposed metrics achieve a significantly higher agreement with human evaluations compared to existing metrics on both video-level and model-level evaluation. Our key contributions are as follows:
\begin{itemize}
\item We propose a comprehensive human evaluation suite, DynamicEval, comprising 100 procedurally curated prompts with diverse camera motions, along with 45k human annotations on 3k videos generated by ten T2V models, designed for video-level evaluation.
\item  In contrast to, VBench baselines that solely rely on deep feature-based metrics and fail to capture fine-grained spatial awareness and long-term temporal context, we propose methods that provide pixel-level grounding.\ 1) For FG subject consistency, we introduce the novel Tracker-FG metric, which measures the temporal smoothness of neighboring tracks within a subject.\ 2) For background scene consistency we introduce MS-Debias to mitigate two key issues (occlusions/disocclusions and foreground objects) that bias motion smoothness in dynamic videos.
\item We conduct extensive experiments on DynamicEval dataset to demonstrate that our proposed metrics achieve stronger agreement with human preferences than baseline metrics (an increase of 0.32 in average PLCC), across both video-level and model-level evaluations. (Fig.~\ref{fig:teaser})


\end{itemize}
\begin{figure*}
  \centering
  \includegraphics[width=0.99\linewidth]{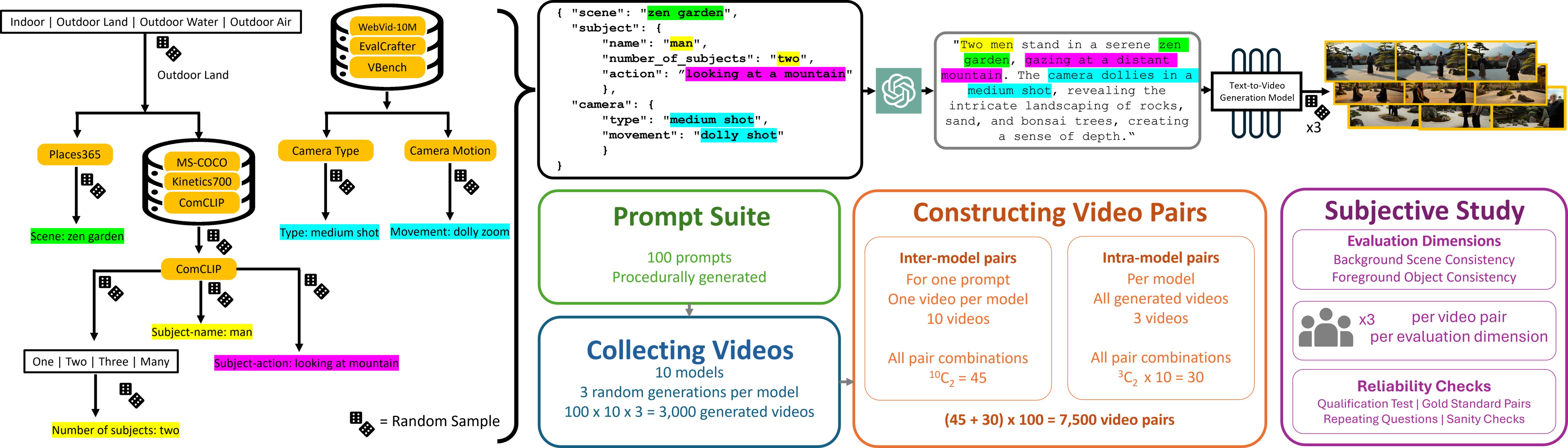}
  \caption{Prompt curation: Scene elements from databases (orange) are sampled into a metadata (JSON format), which GPT-4o converts into descriptive prompts. Dataset: Video pairs generated from a common prompt are annotated via a subjective study.}
  \label{fig:prompt_suite}
\vspace{-5mm}
\end{figure*}

\section{Related Work}
\label{sec:related_works}
\textbf{Text-to-Video Generative Models.}
In the last few years, the field of video generation has experienced a great impetus with diffusion-based generative models~\citep{blattmann2023stable,xing2023dynamicrafter,chen2023videocrafter1,videoworldsimulators2024,bar2024lumiere,polyak2024movie,yang2024cogvideox,LTX-Studio} to generate realistic videos based on textual conditions. In particular, following the seminal works of ~\citet{videoworldsimulators2024, gupta2024photorealistic}, there has been large developments in Diffusion Transformer (DiT)~\citep{peebles2023scalable} based video foundation models both open-source~\citep{opensora, lin2024open, yang2024cogvideox, kong2024hunyuanvideo, wan2025wan, LTX-Studio} and commercial~\citep{runway, luma, minimax, adobe, veo} variants, that can generate long and high-resolution videos. Considering the rapid development and commercialization in this field, it becomes extremely important to develop evaluation criteria and metrics to judge the quality of video foundation models.

\noindent \textbf{Benchmarks and Datasets.}
The availability of the text-to-video models has led to the development of large-scale evaluation benchmarks studies like VBench~\citep{vbench}, VBench++~\citep{vbenchpp}, EvalCrafter~\citep{evalcrafter}, DEVIL~\citep{devil} and GenAIarena~\citep{Genai_arena}. VBench, EvalCrafter and DEVIL provide a large suite for model-level evaluation metrics across several dimensions, with prompts that produce predominantly subject centric and static scenes. Model-level evaluations suffer from aggregation of preferences across all prompts (or videos) in the suite. Different from these benchmark studies and suites, DynamicEval provides a comprehensive suite of pairwise video comparisons that can be used to assess automated metrics specifically for dynamic scenes.

\noindent\textbf{Evaluation Metrics.}
VBench~\citep{vbench} provides a background consistency metric based on the similarity scores between CLIP~\citep{radford2021learning} embeddings of consecutive frames. While it captures the overall content consistency across frames, it fails to detect localized background distortions that require pixel-level analysis. MEt3R~\citep{met3r} introduces a metric based on the computation of 3D point clouds for consecutive frames. Although this metric operates at a fine-grained level, it is vulnerable to errors from inaccurate 3D point estimation by DUSt3R~\citep{dust3r} on generated frames. Vbench also introduces a subject consistency metric that relies on DINO~\citep{dino} feature similarities across consecutive frames. 
DINO models are self-supervised transformer models that are found to attend more to the primary subjects in a frame. However, the reduced resolution of attention maps compared to the original frame size limits their ability to capture fine-grained subject details. Additionally, since they are computed independently at the frame level without leveraging neighboring frames, they are highly sensitive to scene variations, particularly in dynamic videos.
Motivated by these limitations, we propose pixel-level and temporal tracking based methods that move beyond feature-level approaches, enabling fine-grained and interpretable evaluation.


\section{DynamicEval: Dataset}
\label{sec:dataset}
\begin{figure*}
\centering
\begin{subfigure}[t]{0.5\textwidth}
  \centering
  \includegraphics[width=\linewidth]{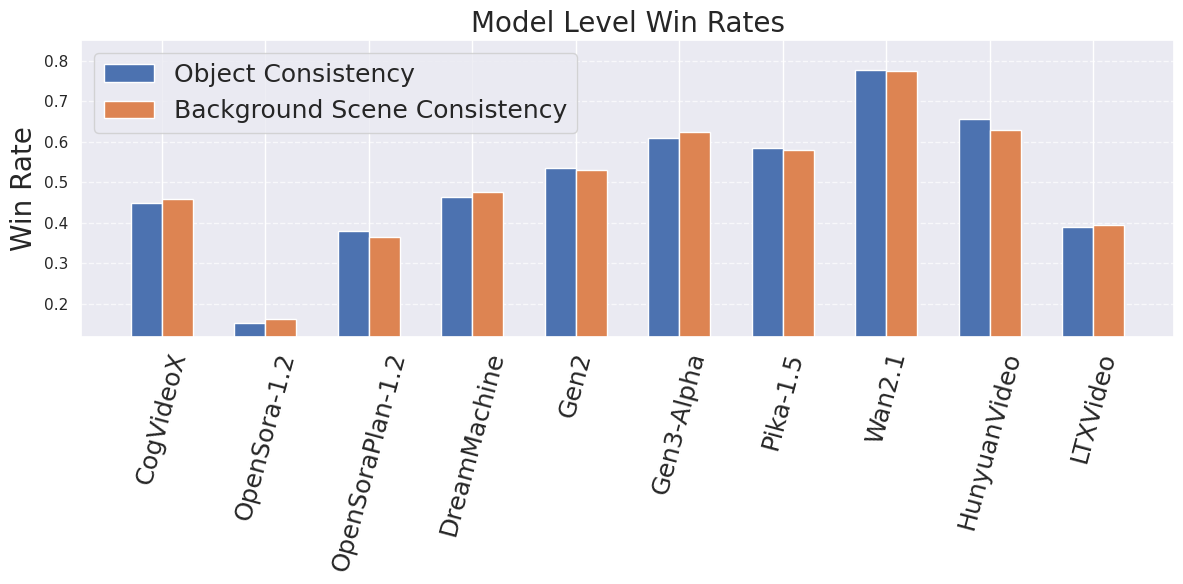}
\hfill
\end{subfigure}
\begin{subfigure}[t]{0.4\textwidth}
  \centering
  \includegraphics[width=\linewidth]{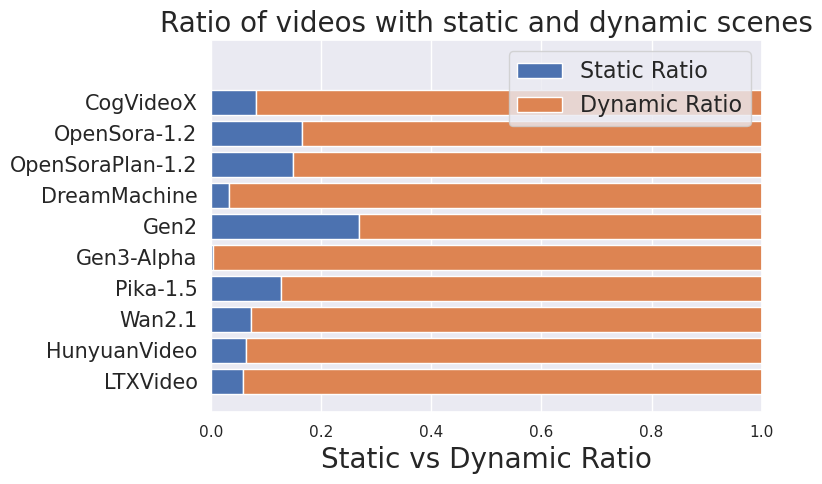}
\end{subfigure}
\vspace{-5mm}
\caption{Dataset analysis: The plot on the \textit{left} shows the average win rates for both evaluation dimensions. The plot on \textit{right} illustrates the percentage of video samples in each model that are static and dynamic.}
\label{fig:dataset_analysis}
\vspace{-5mm}
\end{figure*}

Existing generated video evaluation benchmarks~\citep{vbench, evalcrafter} introduce prompts that are often subject-centric and depict relatively static scenes with little-to-no camera motion. This highlights the need for a prompt suite that targets T2V generation of scenes involving significant camera motion. 
For the purpose of this work, we use the term \textit{`dynamic scenes/videos'} to refer to videos with substantial camera motion.
We introduce DynamicEval, generated video evaluation dataset by carefully curating text prompts that describe diverse camera motions (Ref. Fig.~\ref{fig:dataset_analysis}). Our dataset includes pairwise video preference annotations along two key dimensions of dynamic video quality: 1) foreground subject consistency and 2) background scene consistency.


\subsection{Prompt Curation}

To generate diverse scenes with camera motion, we follow existing works~\citep{heim, vbench, HRS-Bench}, collecting various keywords for scene elements. We randomly sample across three key aspects of dynamic videos: (i) background scene, (ii) primary subject(s), and (iii) camera movement. 
The camera attributes specify a camera type and motion, which we collect from camera-related keywords extracted from video benchmarks~\citep{webvid-10m, evalcrafter, vbench}
(see supplementary for details).
This approach procedurally generates complex scenes that feature highly diverse background settings, subjects, and their motions, as well as camera movements. We construct our benchmark prompt suite of 100 prompts using this pipeline as illustrated in Figure~\ref{fig:prompt_suite}.



\subsection{Subjective Study}
We generate videos using our prompt suite with ten latest state-of-the-art T2V models which includes both open-source (OpenSora~\citep{opensora}, OpenSoraPlan~\citep{lin2024open}, CogVideoX~\citep{yang2024cogvideox}, HunyuanVideo~\citep{kong2024hunyuanvideo}, Wan2.1~\citep{wan2025wan}, and LTXVideo~\citep{LTX-Studio}) and closed-source (DreamMachine~\citep{luma}, Pika~\citep{pika}, Runway Gen2~\citep{gen2}, and Runway Gen3-Alpha~\citep{runway}) models.  
{Among open-source models, we select the largest-parameter variant of each model.}
We collect three videos per prompt per model, totaling 3k videos, making this a comprehensive T2V dataset with both inter-model and intra-model video-level quality comparisons.
We conduct a large-scale subjective study on generated video pairs using Amazon Mechanical Turk (AMT). 
We collect human preferences for foreground subject consistency and background scene consistency. 

\begin{figure*}
  \centering
  \includegraphics[width=0.9\linewidth]{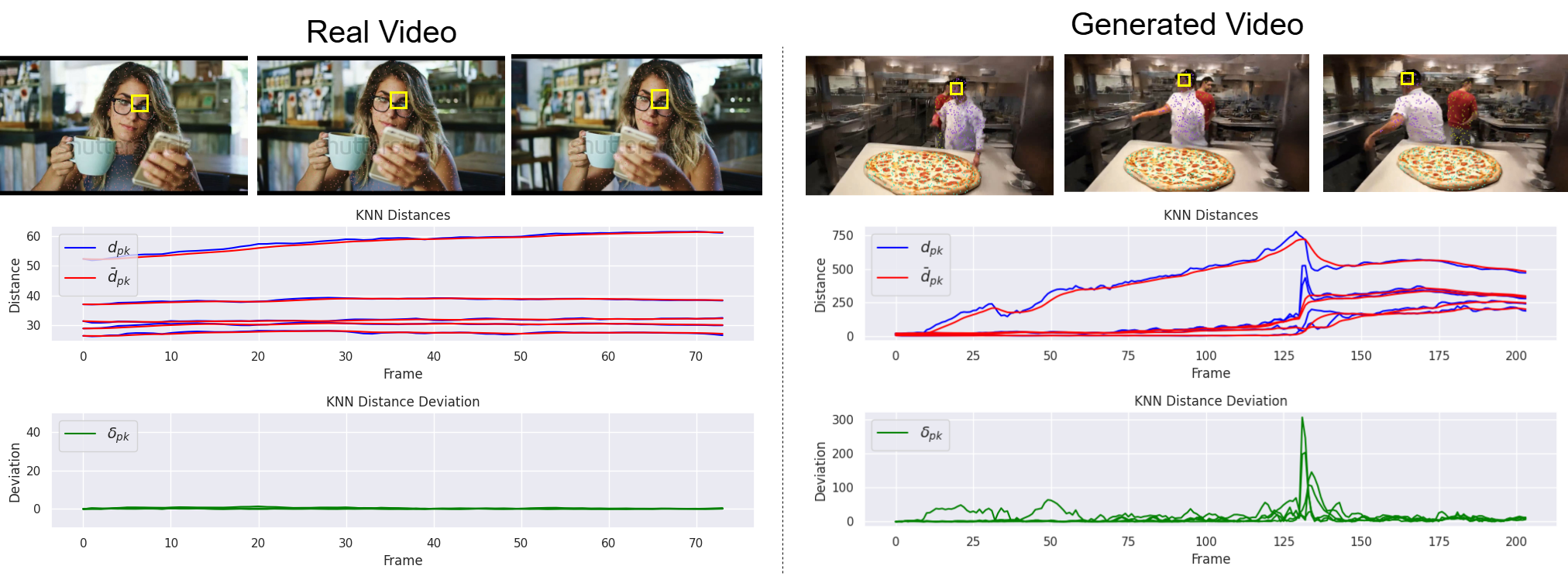}
  \caption{Tracker Motivation: Deviation of neighbor tracks on real video vs generated video with subject distortions. Blue plot: Distance between a nearest neighbor and a candidate point. Red plot: Moving average of blue. Green plot: the deviation between blue and red.}
  \label{fig:FG-Motivation}
\vspace{-3mm}
\end{figure*}

\noindent\textbf{Foreground subject consistency.} Foreground subjects in generated videos may exhibit unnatural shape changes across frames. This evaluation dimension captures how foreground subjects remain consistent throughout the scene. For example, under camera motion, a generative model may fail to preserve the consistency of a human face, as illustrated in the second example of Fig~\ref{fig:FG-Motivation}. 

\noindent\textbf{Background scene consistency.} In generated videos with significant camera movement, the background may undergo unnatural morphing or stretching, leading to localized distortions. We show an example of a generated video demonstrating low background scene consistency in Fig.~\ref{fig:BG_Motivation}.




\noindent\textbf{Crowd-sourced Human Subjective Study.} We conduct a subjective study where we present pairs of videos generated from the same prompt and ask participants to select the preferred video for each evaluation dimension. With $3$ generations each from ten different T2V models, we obtain 30 videos per prompt. Instead of exhaustively collecting all $^{30}C_2$ pairs, we sample 45 inter-model and 30 intra-model pairs. (details in supplementary material), resulting in 7.5k video-pair comparisons per evaluation dimension. To ensure reliable fine-grained annotations we follow standard practices~\citep{live-vqc, konvid} and employ multiple reliability checks, including an initial qualification study, gold standard pairs, repeated questions, and content-related questions. We collect three human annotations per comparison, yielding 
45k fine-grained human annotations.

\subsection{Dataset Analysis}

\textbf{Model-level human preference.} We analyze the average model-level win ratios of videos across inter-model pairs in Fig.~\ref{fig:dataset_analysis} (left). The win ratio calculates the fraction of times a video is selected out of all its comparisons with other videos, with a higher win ratio indicating better model performance. We find that older open-source models generally achieve lower win ratios on both evaluation dimensions, whereas closed-source models generate higher quality videos with better background and foreground subject consistency, verifying the sanity of the study. Interestingly, the latest open-source model, Wan2.1~\citep{wan2025wan} outperforms all other models across both dimensions.

\noindent\textbf{Presence of static scenes in the video generations.} Despite explicitly providing prompts with camera motion descriptions, some models fail to generate dynamic videos. 
To identify static and dynamic scenes, we average the variance of point tracks from CoTracker~\citep{cotracker} across frames to obtain a camera-motion metric. 
Through empirical analysis, we find that the bottom $10\%$ of videos by camera motion values corresponds to static scenes with very little to no camera motion. We treat these as static videos. The ratio of static-vs-dynamic videos per model is shown in Fig.~\ref{fig:dataset_analysis} (right). We find that most open-source models and an older closed-source model (e.g., Runway Gen2) generate many static scenes despite camera motion explicitly mentioned in the prompt. We purposefully retain the static videos in the evaluation to ensure our metrics perform robustly in both static and dynamic scene scenarios.

\begin{figure*}
  \centering
  \includegraphics[width=0.99\linewidth]{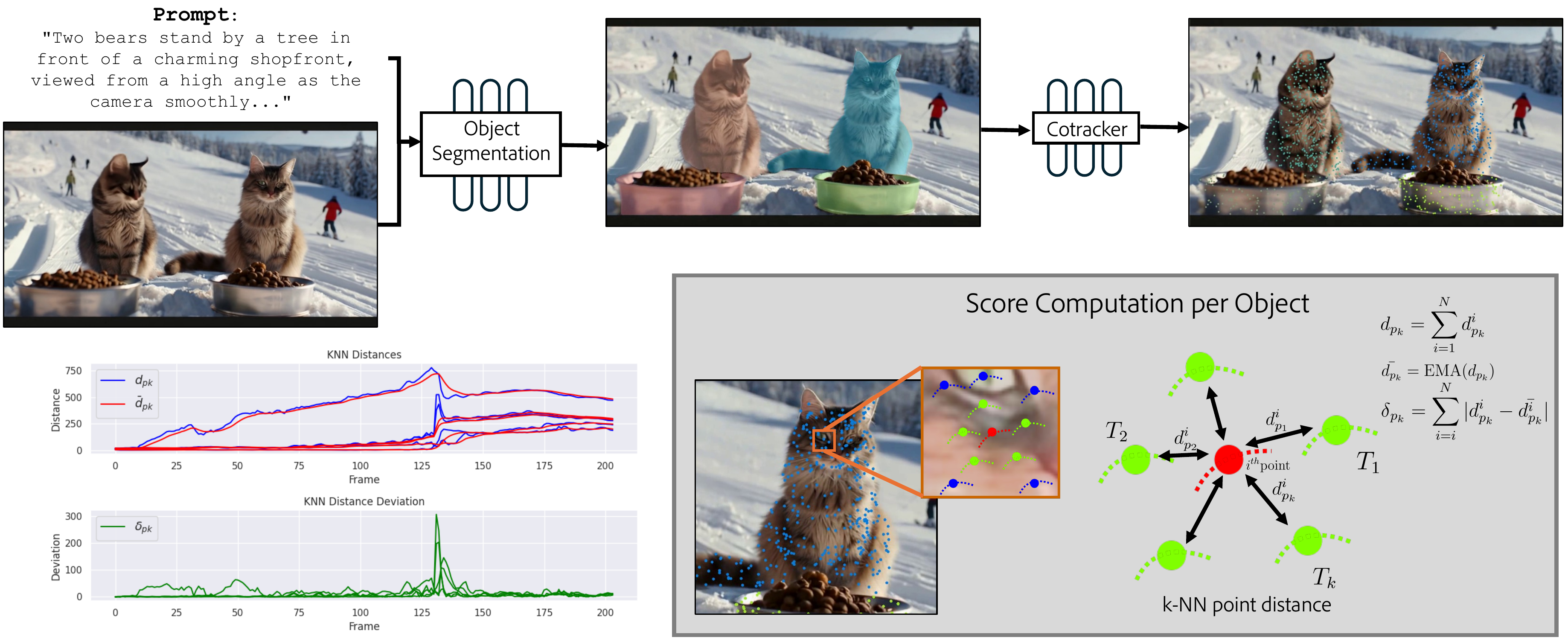}
  \caption{Tracker-FG framework: We track the points inside an object mask across the video. TrackerFG is obtained by computing the average of distances between neighboring tracking points.}
  \label{fig:FG_SAM_Cotracker}
\vspace{-5mm}
\end{figure*}

\section{DynamicEval: Metrics}
\label{sec:metrics}

We focus on the evaluation of two key dimensions of dynamic video quality. For foreground subject consistency, the ideal metric tracks subject shape deformations over time. It should isolate subjects in pixel space and employ point tracking methods for evaluation. In contrast, for background scene consistency, the evaluation should avoid the foreground objects. We now take a closer look at the two dimensions and introduce our new metrics below.

\begin{figure*}
  \centering
  \includegraphics[width=0.94\linewidth]{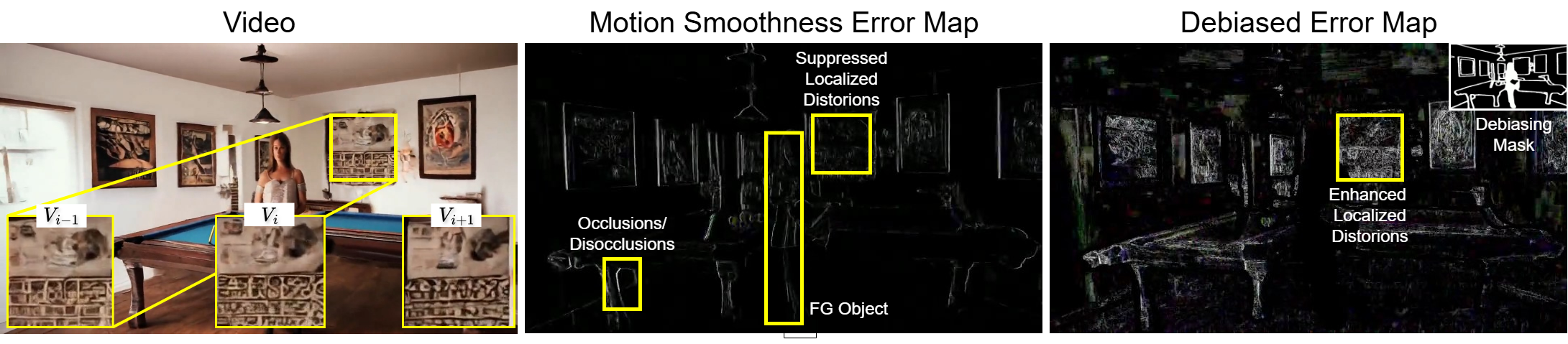}
  \caption{Motion Smoothness error maps: The zoomed-in regions show localized distortions visible across frames. VB-MS shows large errors near edges and foreground objects, suppressing the localized distortions. After debiasing, the localized distortions are visible.} 
  \label{fig:BG_Motivation}
\vspace{-5mm}
\end{figure*}

\subsection{Foreground Subject Consistency}
\label{sec:vb-fg}
In our analysis, we find that the only existing metric that actually focuses on the foreground consistency is the VBench subject consistency (VB-SC) metric~\citep{vbench}. VB-SC computes the similarities of the DINO~\citep{dino} characteristics in consecutive frames. These features are computed at the frame-level and rely on attention maps that are much smaller than the frame sizes, making them ill-suited for evaluating the temporal aspects of subject consistency in finer detail.  
To address this we use point tracks, which capture fine-grained details and long-term temporal context. 
In Figure \ref{fig:FG-Motivation} we observe low neighbor track deviation in real videos, while with inconsistent subjects in generated video, the deviation is high. Motivated by this, we propose Tracker-FG as shown in Figure \ref{fig:FG_SAM_Cotracker}.
To detect foreground subjects, we leverage the prompt that was used to generate the video. 
We use an LLM to identify subjects in prompt and pass the list of subject names into the GroundingDINO~\citep{GroundingDINO} model to localize these subjects in the video. 
The localizations are further transformed into masks and propagated across all frames using SAM2~\citep{sam2024vid} to obtain the video segmentation mask for each subject as $\objectmask_{i\oid}; \oid \in \{1, 2, \cdots, \totalobjects\}$, where $\totalobjects$ denotes the number of subjects in the scene, and $i \in \{1, 2, \cdots, \totalframes\}$ refers to the frame number. The individual subject masks are merged into final mask, $\objectmask_i$, using a union operation.

We use a state-of-the-art point tracking model, CoTracker~\citep{cotracker}, to track randomly sampled points inside the subject masks. 
While designing the evaluation metric using these tracks, we ensure two major requirements. First, the metric should evaluate the consistency of neighboring tracks; second, it should remain invariant to global subject motion and camera movement. For $\totalobjects$ subjects in the video, let all the point tracks be denoted as $\pointtrack_\pid^\oid; \pid \in \{1, 2, \cdots, \totalpoints_\oid\}$, where $\totalpoints_\oid$ corresponds to the total number of points tracked in the $\oid^{\text{th}}$ subject. For each $\pointtrack_\pid^\oid$, we first identify the k-nearest neighbor tracks of the point as $\pointtrack_k^\oid \in \{\text{k-NN}(\pointtrack_\pid^\oid)\}$ from each frame. We find the distance to nearest neighbors, $\trackdist_{\pid k}^\oid$, as, $\norm{\pointtrack_\pid^\oid - \pointtrack_k^\oid}_2$.

By computing distances between neighboring points, the metric is invariant to global subject motion and camera movement. Finally, 
we compute the moving average of $\trackdist_{\pid k}^\oid$ across frames and find the mean absolute error (MAE) of $\trackdist_{\pid k}^\oid$ with its moving average (EMA) track $\bar{\trackdist}_{\pid k}^\oid$ as $\diffdist_{\pid k}^n$. This reveals the noisy trajectories in the neighboring tracks, capturing the distortions in each subject. To obtain the subject consistency error, we average $\diffdist_{\pid k}^n$ across the $k$ neighbours of each tracker and across all tracking points $\totalpoints_\oid$.  Finally, we take an average of subject scores to obtain subject inconsistency score, which is then subtracted from 1 to obtain the final Tracker-FG metric.

\subsection{Background (BG) Consistency}
\label{sec:vb-bg}
In this section, we first review the commonly used evaluation metrics introduced in VBench~\citep{vbench}, EvalCrafter~\citep{evalcrafter}, and Met3R~\citep{met3r}. These metrics have been shown to perform well on existing prompt suites, which are predominantly subject-oriented and involve very little (or static) camera motion. To assess their limitations, we evaluate them on our database to analyze how they behave when applied to videos with significant camera motion.

\noindent\textbf{Background consistency metrics.}
We evaluate the baseline metrics on the background scene consistency dimension of our DynamicEval dataset by computing the pairwise video preference of each method with respect to subjective human preferences. From VBench~\citep{vbench}, we evaluate background consistency (VB-BG), flickering (VB-flickering), and motion smoothness (VB-MS), and from EvalCrafter~\citep{evalcrafter}, we include the semantic consistency (EC-semantic) metric. In addition, we also evaluate Met3R~\citep{met3r}. 
The evaluation results presented in Table \ref{tab:BG_analysis}shows that VB-BG, VB-MS, and Met3R perform decently among the evaluated metrics. Due to its simplicity and access to pixel-level evaluation, we analyze VB-MS for background consistency.

\begin{table}
  \caption{Pairwise video selection accuracy of existing metrics on BG scene consistency.}
  \label{tab:BG_analysis}
  \centering
  \begin{adjustbox}{max width=0.47\textwidth}
      
  \begin{tabular}{l|lllll}
    \toprule
    Metrics & VB-BG & VB-flickering & VB-MS & EC-semantic & Met3R\\
    \midrule
    Accuracy & 56.0 & 51.3 & 53.7  & 52.0 & 54.9  \\    
    \bottomrule
  \end{tabular}
  \end{adjustbox}
  
\vspace{-5mm}
\end{table}

\noindent\textbf{Analysis of VB-MS.}
As VB-MS is a pixel level metric, it captures localized distortions in the background very well. 
Motion smoothness uses the optical flow model, RAFT~\cite{raft}, to predict an intermediate frame between two alternate frames. The absolute difference between the predicted frame and the original frame provides an error map, which is spatially and temporally averaged to obtain the inconsistency score.
To understand how the motion smoothness captures localized distortions, we analyze the error map of a generated video in Figure \ref{fig:BG_Motivation}. 
We also observe that although the error map captures localized issues, it is highly influenced by regions near object edges. This behavior primarily arises due to object occlusions with moving camera. 
Further, the foreground subjects also significantly contribute to the error, as subject motion is harder to predict with optical flow, and does not reflect background consistency. We leverage these observations and discuss techniques to debias and improve motion smoothness below.

\begin{figure*}
  \centering
  \includegraphics[width=0.9\linewidth]{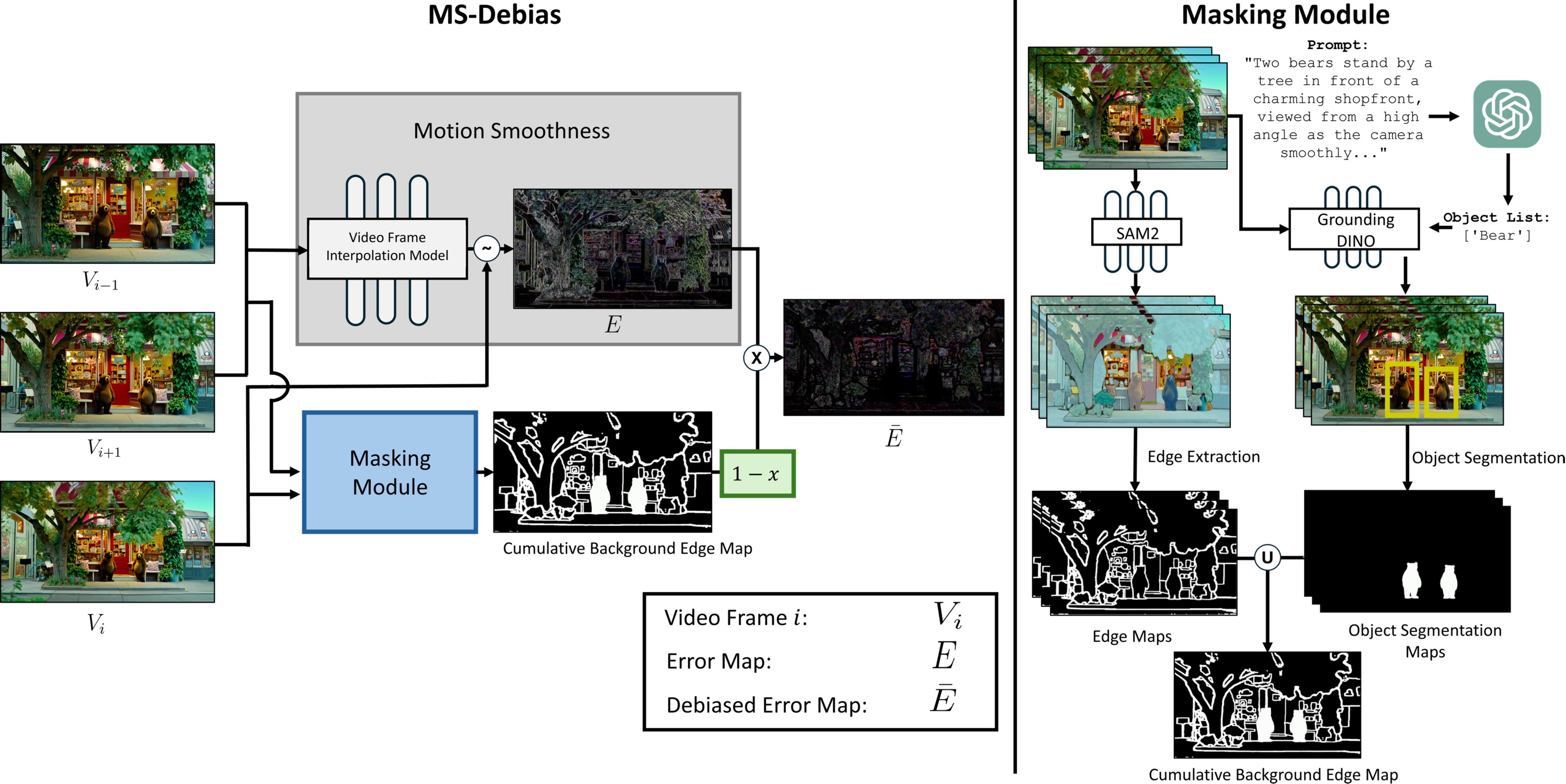}
  \caption{MS-Debias: obtains debiased motion smoothness error maps by masking out foreground subjects and occlusions. It is applied at multiple scales with Gaussian pyramid downsampling, for improving the robustness of the metric.} 
  \label{fig:MS_debias}
\vspace{-3mm}
\end{figure*}

\begin{table*}
    \caption{Performance Comparisons: The first table shows pairwise preference accuracy on the DynamicEval database, and the second table evaluates the Top-k video selection accuracy. On the full dataset, our method, MS-Debias outperforms the VB-BG by $2.2$\% points and on the full agreement subset by $2.4$\% points. Similar boosts can be seen for subject consistency pairwise accuracy.}
    \label{tab:maincomparison}
    \centering
    \begin{tabular}{cc}
        \begin{adjustbox}{max width=0.43\textwidth}
        \begin{tabular}{lcc}
            \toprule
            \multicolumn{3}{c}{Subject Consistency (Pairwise Acc.)} \\
            \midrule
            Method & Full Dataset & Full Agreement \\
            \midrule
            VB-SC & 56.2 & 58.8 \\
            Tracker-FG (Ours) & \textbf{58.2} & \textbf{62.7} \\
            \midrule
            \multicolumn{3}{c}{Background Consistency (Pairwise Acc.)} \\
            \midrule
            VB-BG & 56.0 & 59.3 \\
            VB-MS & 53.7 & 56.8 \\
            MS-Debias (Ours) & \textbf{58.2} & \textbf{62.7} \\
            \bottomrule
        \end{tabular}
        \end{adjustbox}
        & \hspace{6mm}
        \begin{adjustbox}{max width=0.48\textwidth}
        \begin{tabular}{lccccc}
            \toprule
            \multicolumn{6}{c}{Subject Consistency (Top-k)} \\
            \midrule
            Method & Top-1 & Top-2 & Top-3 & Top-4 & Top-5 \\
            \midrule
            VB-SC & 16.2 & 29.1 & 39.5 & 50.0 & 59.3 \\
            Tracker-FG (Ours) & \textbf{19.7} & \textbf{31.4} & \textbf{41.9} & \textbf{55.8} & \textbf{62.8} \\
            \midrule
            \multicolumn{6}{c}{Background Consistency (Top-k)} \\
            \midrule
            Method & Top-1 & Top-2 & Top-3 & Top-4 & Top-5 \\
            \midrule
            VB-BG & 10.1 & 31.5 & 41.6 & 55.1 & 65.2 \\
            MS-Debias (Ours) & \textbf{14.6} & \textbf{34.8} & \textbf{44.9} & \textbf{57.3} & \textbf{69.7} \\
            \bottomrule
        \end{tabular}
        \end{adjustbox}
    \end{tabular}
\end{table*}

\noindent\textbf{Debiasing motion smoothness metric.}
To address the camera motion and foreground subject bias in motion smoothness, we carefully construct masks around object edges
as shown in Figure \ref{fig:MS_debias}. 
To detect object boundaries, we employ the auto-object detector of SAM-2~\citep{sam2024vid} and propagate the object segmentation masks across all the frames. The segmentation masks are then converted into edge boundaries by applying a morphological gradient, and thickened by dilation, to obtain the final edge map, $\edgemap_i$, where $i \in \{1, 2, \cdots, \totalframes\}$ refers to the frame number.
We leverage the subject masks $\objectmask_{i}$ used earlier in Section~\ref{sec:vb-fg}. 
The final debiased error map, $\bar{\errormap_i}$, is computed as, 
\begin{equation}
    \bar{\errormap_i} = \errormap_i \times \left(1 -\left(\edgemap_i \cup \objectmask_i \right)\right)
\end{equation} 
where $\errormap_i$ is the error map obtained from VB-MS for the $i^\text{th}$ frame.
We note that our debiased error maps are devoid of the localized issues as shown in Figures \ref{fig:BG_Motivation}. 
Additionally, to mimic the multi-scale processing capabilities of the human visual systems~\citep{mssim, strred, vmaf}, we apply our debiased motion smoothness to multiple downscaled versions of videos and aggregate the scores to obtain the final metric.

\begin{table*}
  \caption{Video-Pair Configurations: Performance of each metric when the pair contains different configurations of static-dynamic pairs. The improved metrics ensure no bias, for static videos over the dynamic ones, in the proposed Tracker-FG and MS-Debias metrics.}
   \vspace{-2mm}
    \centering

    \begin{tabular}{lcc|lcc}
        \toprule
        \multicolumn{3}{c}{Subject Consistency} & \multicolumn{3}{c}{Background Consistency} \\
        \midrule
        Method & static-dynamic & dynamic-dynamic & Method & static-dynamic & dynamic-dynamic \\
        \midrule
        VB-SC & 57.1 & 57.3 & VB-BG & 55.2 & 56.1 \\
        Tracker-FG & \textbf{60.3} & \textbf{58.1} & MS-Debias & \textbf{58.0} & \textbf{56.7} \\
        \bottomrule
    \end{tabular}
  \label{tab:perf_static}
\vspace{-5mm}
\end{table*}

\section{Experiments and Results}
\label{sec:experiments}

We evaluate the proposed metrics on the foreground subject consistency and background scene consistency dimensions of our DynamicEval benchmark. 
We evaluate our models on both the full dataset and a subset where all annotators agree unanimously on their selection (full agreement).

\noindent \textbf{Video-level comparisons.}
To compare video-level performance, we evaluate the methods on two metrics. Pairwise preference accuracy: The proportion of times a metric selects the same video as humans. Top-k video selection accuracy: The proportion of times the ground-truth best video is ranked within the top-k predictions of the metric. For Top-k evaluation, we obtain the ground-truth ranking of videos given a prompt through win-ratios (see supplementary for details).
We compare our metrics with the baseline metrics proposed in VBench in Table \ref{tab:maincomparison}. 
For the foreground subject consistency dimension of DynamicEval database we evaluate VBench subject consistency (VB-SC) and our Tracker-FG metric.
The final pairwise scores are computed as an average of both the VB-SC and our Tracker-FG to incorporate the best of both feature consistency and tracker-level consistency.
Our proposed metric outperforms the baseline, highlighting the advantage of explicitly focusing on the foreground subjects and computing motion invariant metrics. 
For the background consistency dimension of DynamicEval database, we evaluate VB-BG, VB-MS and our debiased motion smoothness metric (MS-Debias). 
Similar to foreground consistency, we also average our metric and VB-BG to combine the strengths of feature and pixel-level consistency. 
Our final metric outperforms both baselines, highlighting the effectiveness of debiasing and multi-scale processing in video-level evaluation.
Additionally, we qualitatively analyze these metrics (Suppl. Sec. D.2, and E.2.2).

\noindent\textbf{Model-level comparisons.}
Similar to the model-level comparisons presented in VBench, we evaluate the ability of our metrics to rank the models. To obtain a model-level ground truth score, we compute the average of video win ratios across each model. To evaluate the correlation of model-level ground truth scores with metric scores, we compute their Pearson's Linear Correlation Coefficient (PLCC) in Table \ref{tab:model-level}. Both Tracker-FG and MS-Debias outperform the baseline by a significant margin in terms of correlation with human judgments for selecting the best model.

\begin{table}[t]
    \caption{PLCC of model-level win ratios between the metrics and human preference. We evaluate these on the full agreement subset.}
    \centering
    \begin{tabular}{lc|lc}
        \toprule
        Method (FG) & PLCC & Method (BG) & PLCC \\
        \midrule
        VB-SC & 0.334 & VB-BG & 0.551 \\
        Tracker-FG & \textbf{0.772} & MS-Debias & \textbf{0.743} \\
        \bottomrule
    \end{tabular}
    \label{tab:model-level}
\vspace{-3mm}
\end{table}

\noindent\textbf{Pairwise preference on static and dynamic scenes.}
Although our prompt suite is designed to generate dynamic scenes, a fraction of the generated videos are static, as shown in Figure \ref{fig:dataset_analysis} (right). 
In a pairwise comparison, there are three scenarios: both videos are static (static-static), one of them is static (static-dynamic), and both are dynamic (dynamic-dynamic). As the occurrence of static-static pairs is very low, we omit evaluation on this subset. We evaluate the preference performance on each of these subsets in Table \ref{tab:perf_static}. The experimental results show that our proposed metrics consistently outperform the baselines when dynamic scenes are present.

\noindent\textbf{Impact on different types of camera motion.}
We broadly classify camera motion described in each prompt into four subsets: linear translational motion (dolly shot, tracking shot, slider move), curved translational motion (arc shot, panning around subject), handheld motion and rotational motion (camera turns left/right, pedestal shot). We then compute the performance of each subset individually as shown in the Table~\ref{tab:cam-motion}.
FGTracker benefits the most from videos with handheld motion as the neighbor distances of trackers handle the subject consistency well.
\begin{table}[t]
  \caption{Pairwise preference accuracy on types of camera motion.}
  \vspace{-1mm}
  \label{tab:cam-motion}
  \centering
  \begin{adjustbox}{max width=0.47\textwidth}
  \begin{tabular}{lcccccc}
  \toprule
    Method & Full\;\; & Linear\; & Curved\; & Handheld & Rotatory\\
    \midrule
    \multicolumn{6}{c}{Foreground Subject Consistency}\\
    \midrule
    VB-SC      & 56.2 &55.9 &57.2 &54.5 &56.1\\    
    Tracker-FG & \textbf{58.2} &\textbf{57.6} & \textbf{60.5} &\textbf{59.7} &\textbf{56.5}\\      
    \midrule
    \multicolumn{6}{c}{Background Scene Consistency}\\
    \midrule
    VB-BG & 56.0 & 55.2 & 54.1 & 57.9 &57.1\\    
    MS-Debias & \textbf{58.2} &\textbf{57.7} &\textbf{56.5} &\textbf{61.2} &\textbf{57.9}\\     
    \bottomrule
  \end{tabular}
  \end{adjustbox}
\vspace{-5mm}
\end{table}
We see larger gains in MS-Debias background consistency with linear and handheld motion as these introduce more parallax. We provide  more visualizations and analysis in supplementary.

\section{Discussion}
\label{sec:limitations}
\noindent \textbf{Limitations.}\ The practical use of our proposed metrics relies on off-the-shelf vision models for optical flow, point tracking, and segmentation.
Large-scale pre-training enables these vision models to capture visual primitives (edges, textures, objects, semantics) that transfer across domains. Notably, RAFT and CoTracker, despite being trained on synthetic 3D scenes, generalize well to out-of-distribution real videos, and naturally to generated videos as well.
Prior state-of-the-art works in evaluation metrics for T2V~\citep{vbench, evalcrafter, vbenchpp, met3r} further show that such models remain effective for evaluating generated videos. Due to the plug-and-play nature of our metrics, we can keep improving them in the future with improved base models.

In summary, we demonstrate that the existing evaluation frameworks for T2V models have a blind spot for camera motion. To mitigate this, we propose a large-scale and thoughtful benchmark, DynamicEval, to augment the existing ones. As existing metrics fail on DynamicEval, we introduce two novel metrics: Tracker-FG and MS-Debias to robustly evaluate the subject and background consistency, respectively, for videos with camera motion. These metrics significantly improve over the baselines and display a significantly high correlation with human preferences without being trained on any human labels. 
This makes the proposed metrics, to be promising candidates for incorporation into preference-based training methods (\ie RL) to further improve model performance in future research.

\section*{Acknowledgments}
This work was supported in part by the Kotak IISc AI–ML Centre
(KIAC).

{
    \small
    \bibliographystyle{ieeenat_fullname}
    \bibliography{references}
}

\clearpage
\maketitlesupplementary
\appendix

\section{Visual Examples}
Please note that we have provided an HTML page containing different visualizations supporting our results. Extract the zip file and run \texttt{SupplementaryVideos.html}. We will refer to some of these visualizations in the subsequent sections.

\section{Dataset curation and Subjective Study}
\subsection{Prompt Curation}
The prompts are generated based on three key aspects: background scene, object/subject, and camera attributes.
We collect various classes of these key aspects from existing databases to generate the text prompts. 
We describe the details of this data collection:
\begin{enumerate}
    \item \textbf{Background Scene:} We use the Places365 \cite{places365} dataset to obtain a list of 434 background scenes and manually classify them into indoor, outdoor-land and outdoor-water. This classification helps pair the other key aspects in a realistic manner with the relevant background scenes.
    \item \textbf{Object/Subject:} To have a primary object/subject of focus, we collect 80 categories of objects described in MS-COCO dataset \cite{ms-coco} in which 19 are human/animal subjects while others are inanimate objects. Further, we collect 278 human/animal/vehicle subjects from ComCLIP dataset \cite{comclip}.
    \begin{enumerate}
        \item \textbf{Subject Action:} From Kinetics700 \cite{kinetics700} we first follow a graph based algorithm \cite{graph_algo} 
        to select 100 semantically diverse human actions and pair them with human subjects. The semantic diversity is achieved by clustering the sentence embeddings of the human action phrases and equally sampling from each cluster. ComCLIP \cite{comclip} also provides human/animal/vehicle actions that are paired with relevant subject categories along with the scene classification (indoor/outdoor).
        \item \textbf{Number of Subjects:} Many generative models fail to keep the subject shape consistent if there are more than one subjects in the scene, causing merging/splitting of subjects. To enable evaluation of such issues we add a subject attribute that specifies the number of subjects in the scene (`one', `two', `three' or `many').
    \end{enumerate}
    \item \textbf{Camera Attributes:} The crucial part of our benchmark is to have camera motion that helps evaluate the model's ability to generate good quality dynamic videos. There is no existing work that explicitly provides a list of camera types and motions. Thus, we extract such keywords from prompt benchmark datasets \cite{webvid-10m, evalcrafter, vbench} manually and classify them into camera type and motion often paired together and classified into outdoor/indoor. 
    \begin{enumerate}
        \item \textbf{Camera Type:} The camera type describes the initial camera setting while capturing a video. It could describe the focal length, positioning or type of the camera (wide angle, medium shot, aerial shot, low angle shot or helicopter camera). We collect 15 camera types from the prompt benchmarks.
        \item \textbf{Camera Motion:} To introduce camera motion, we collect a list of 26 diverse camera motions (smooth dolly move, arc shot, trucking shot or follow subject).
    \end{enumerate}
\end{enumerate}
We randomly sample each element and its paired attributes to construct a scene metadata in JSON format and ask GPT-4o to generate a plausible description of the scene. An example of generated prompt from such metadata is shown in Tab.~\ref{tab:prompt_gen}.

\begin{table*}
\caption{Structured metadata (left) and its descriptive narrative (right).}
\label{tab:prompt_gen}
\centering
\begin{tabular}{c|c}

\toprule
Scene meta-data & Generated prompt\\
\midrule
\begin{minipage}[t]{0.55\textwidth}
\begin{verbatim}

{
  "setting": "indoor",
  "action_dataset": "comclip",
  "metadata": {
    "scene": "auto factory",
    "subject": {
      "name": "dog",
      "number_of_subjects": "one",
      "action": "drinking the water"
    },
    "camera": {
      "type": "ground shot",
      "movement": "dolly shot"
    },
    "extra attibutes": ""
  },
}

\end{verbatim}

\end{minipage}
&
\begin{minipage}[t]{0.35\textwidth}
\centering
\vspace{1.8cm}
"In an auto factory, a lone dog drinks water from a puddle amidst hulking machinery and assembly lines. The ground shot dolly camera moves forward, revealing industrial surroundings with the dog in sharp focus."
\end{minipage}
\\
\bottomrule
\end{tabular}

\end{table*}

\subsection{Examples for each evaluation dimensions}

We have provided examples of video pairs and their ground truth scores in ``Pairwise Comparisons" section in our HTML page. For convenience we have arranged the video pairs such that the video on the left has higher quality. Most commonly background scene consistency gets affected by localized distortions in the background scene, and for foreground object consistency, humans prefer rigid objects over morphing.

\subsection{Video Generation and Constructing Pairs}
We have provided a grid of generated videos for someof the prompts in "Dataset Videos" section in the HTML page. The video grid is arranged such that each column represents a different model and each row in the column contains different random generations from the model. Most of the videos contain significant camera motion.

We conduct a subjective study where we provide a pair of videos generated from the same prompt and ask the human subjects to select the video based on each evaluation dimension. Given a prompt that has $10 \times 3 = 30$ videos, we generate a controlled number of pairs that compare videos generated from the same model and across models, instead of evaluating on all $^{30}C_2$ pairs.
All combinations of pairs from the same model are compared, generating $^3C_2=3$ pairs per model per prompt, $^3C_2\times 10\times 100=3,000$ intra-model pairs. For inter-model comparisons, one video is randomly selected from each model. All combinations of pairs across the selected videos yield $^{10}C_2=45$ inter-model pairs per prompt and $^{10}C_2\times 100=4,500$ inter-model pairs in total. Combining both, we evaluate $3,000+4,500=7,500$ video pairs in total. 

Each video pair is evaluated across the $2$ dimensions. Thus, there are $((2\times100)\times(30+45)=15,000$ evaluation pairs in total. We collect 3 subjective ratings per pair totaling $3\times15,000=45,000$ human ratings. We employ various levels of reliability checks to ensure the quality of annotated data. 

\subsection{Subjective Study}
\textbf{Qualification Test:} First, a pool of workers are presented with a qualification test in which they are provided an instruction video that outlines the details of our study with examples shown for each dimension. The qualification test checks if the workers understand the instructions by asking 10 multiple choice questions (MCQs). Further, we provide 3 gold standard video pairs and ask the workers to select the correct video from each pair for specific evaluation dimensions. The workers are qualified for the main study if they obtain a qualification test score greater than 8 out of 10 and select the correct video in all the 3 pairs. Additionally, we filter the qualified workers on their study setup (screen resolution, size, device). Through the qualification test we select 65/200 workers for the main study.

\textbf{Main Subjective Study:} The main subjective study is designed with more internal reliability checks. In each Human Intelligence Task (HIT), we collect ratings on 15 video pairs, and each video pair is evaluated on all dimensions, totaling 45 pair evaluations on average. We employ 3 new reliability checks involving repeated questions, gold standard pairs and video level sanity checks. For repeated questions, we select 2 out of 15 pairs and repeat them in the HIT after shuffling the video pair and the evaluation dimensions. We manually collect a gold standard set that contains annotations for pairs that are easy to differentiate. We add two random 2 gold standard pairs that contains 2-3 questions. Finally, we manually collect a set of sanity check questions (MCQs) about the content in a video pair; per HIT, we include 1 sanity check that contains two questions. Thus, there are $15+(2+2+1)=15+5^{\text{ReliabilityChecks}}=20$ video pairs in total in one HIT. In the 5 video pairs used for reliability checks, there are around 12 to 14 questions. The HITs are approved if the worker clears atleast 80\% of the reliability questions. With all these checks in place, we collect a highly reliable set of annotations for all the video pairs.

\textbf{Worker Compensation:} The worker compensation is fixed based on the US federal laws for a minimum wage of \$7.5 per hour. The qualification test takes around 8 minutes. Setting \$7.5 per hour, each approved qualification test is compensated with \$1. In the main study, one HIT takes around 35 minutes. Setting a wage of \$8 per hour, each approved worker is compensated with \$5.

\begin{figure*}
  \centering
  \includegraphics[width=0.99\linewidth]{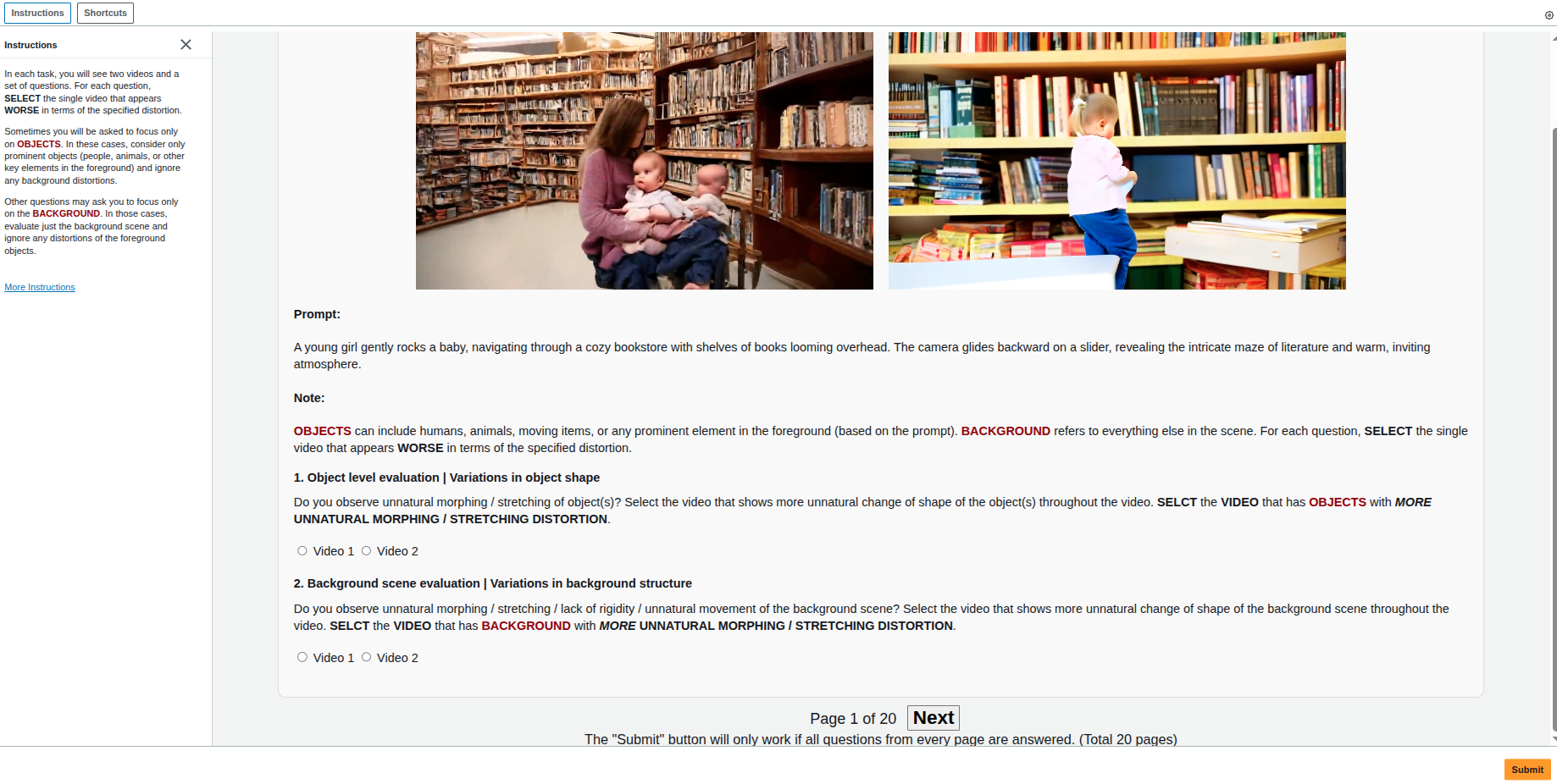}
  \caption{A screenshot of the User Interface for pairwise comparison} 
  \label{fig:UI_Screenshot}
\end{figure*}

\subsubsection{User Interface:}
We design a simple user interface for an HIT with multiple pages. Each page contains a video pair and two questions (evaluation dimensions) as shown in Fig.~\ref{fig:UI_Screenshot}. Both videos are set to autoplay on repeat for convenience. The users can make them full screen if required. We provide short instructions on the left and a separate page with detailed instructions, which contains an instruction video used in the qualification study, with examples detailing how to select a video for each dimension. The human subject must select the video with more distortion with respect to the evaluation dimension. Even if both videos look equally distorted, the subjects are prompted to re-watch and find subtle differences. The workers are allowed to submit the annotations only if all questions from all pairs are answered in the HIT.

\subsection{Foreground object consistency on video pairs without objects}
\label{sec:FG-GT-Construction}
Despite providing foreground object descriptions, some T2V models (mostly older or smaller models) still fail to generate scenes that include the specified objects. We conducted a manual check and observed that out of all the generated videos, around 4\% of the videos fail to show the primary objects. Further, after constructing video pairs, only 1\% of the pairs contained both videos without object generation, while in 6\% of the pairs, one of the videos failed to generate objects. We discard the 1\% cases where both videos fail to generate foreground objects. In the 6\% of the cases where one video failed to generate the object, that particular video is automatically preferred less, as the model was incapable of generating the foreground object. For the subjective study, we only provide the remaining 94\% of the pairs for human evaluation.

\begin{figure*}
\centering
\begin{subfigure}[t]{0.69\textwidth}
  \centering
  \includegraphics[width=\linewidth]{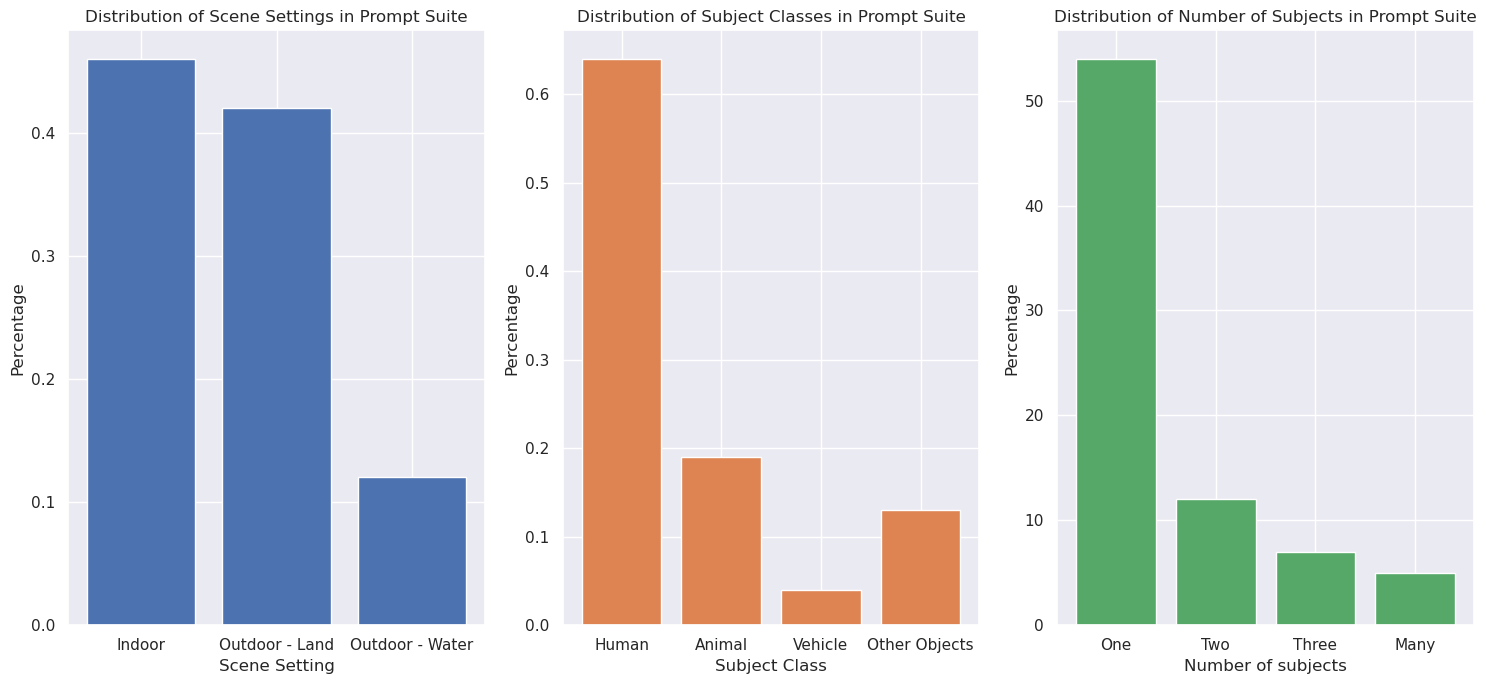}
  \caption{}
  \label{fig:dataset_distribution}
\end{subfigure}
\begin{subfigure}[t]{0.3\textwidth}
  \centering
  \includegraphics[width=\linewidth]{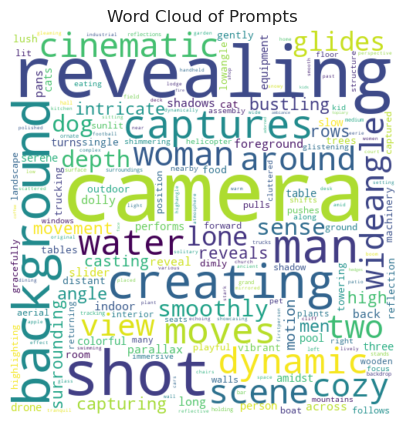}
  \caption{}
  \label{fig:wordcloud}
\end{subfigure}
\caption{Dataset analysis: Percentage of categories of scene elements in the prompts suite, and the prompt wordcloud. A majority of the words focus towards camera movement.}
\end{figure*}

\subsection{Inter-Annotator Agreement}
To verify that the collected preferences are reliable enough to serve as ground truth, we measure inter-annotator agreement separately for the two pairwise dimensions, foreground consistency and background consistency. Each video pair receives a forced binary choice from three independent reliable annotators (after approval filtering), so we report Krippendorff's $\alpha$ (nominal), which is well suited to a design with a variable number of raters per item, as our primary metric, and complement it with Fleiss' $\kappa$ computed over the subset of pairs with exactly three votes and the raw pairwise percent agreement. In addition, every HIT embeds a small number of repeated pairs with the two videos shown in swapped order, which lets us measure each annotator's self-consistency.

Across the foreground and background pairs, we obtain $\alpha = 0.34$ / Fleiss' $\kappa = 0.34$ for foreground consistency and $\alpha = 0.30$ / Fleiss' $\kappa = 0.30$ for background consistency, with pairwise agreement of $0.67$ and $0.65$ respectively. These values fall in the ``fair'' range of the Landis–Koch scale, which is expected for fine-grained subjective comparisons: only about half of the pairs are judged unanimously, reflecting the large fraction of genuinely close comparisons where a binary forced choice mechanically lowers chance-corrected agreement. Crucially, intra-annotator consistency on the swapped-repeat pairs is very high ($0.98$ and $0.98$), indicating that individual annotators are highly self-consistent and that the moderate agreement stems from legitimate between-annotator subjectivity on difficult pairs rather than from noisy labeling. We therefore aggregate the three judgments per pair by majority vote, which yields a stable ground-truth signal despite the inherent subjectivity of the task.

\subsection{Additional data collection}
Adding to the two key aspects of background scene consistency and foreground object/subject consistency, we have collected data on more evaluation dimensions for future work:
\begin{enumerate}
    \item \textbf{Foreground object/subject consistency:} A poorly generated video can have foreground objects unnatural changes in their shape and size across the video, irrespective of the background scene being consistent. The object level evaluation can be divided into three dimensions, namely:
    \begin{enumerate}
        \item \textbf{Object shape variation:} This dimension evaluates unnatural morphing, stretching, merging or splitting of foreground objects/subjects in a scene. For example, under camera motion, a generative model may fail to keep the shape of a table in a scene rigid.
        \item \textbf{Relative object size variation:} In some cases, even if the shape of the object is consistent across frames, the relative size of the object with respect to the background scene may change unnaturally when there is a camera motion. For example, the generative model fails to keep the rate of change of object size proportional to the speed of camera movement. 
        \item \textbf{Unnatural object size:} The generative model can also fail to capture the real life proportions of different objects in a scene, irrespective of objects being consistent in shape or size across the video. One may be able to assess the size unnaturalness at a frame level, however, unless the camera moves, one cannot position an object in a scene to be closer or farther from the camera. Thus, this dimension must be evaluated at a video level.
    \end{enumerate}
    \item \textbf{Background scene evaluation:} Regardless of whether the objects in the video have consistent shape, the background scene can have unnatural variations affecting the perceptual quality. The background consistency aspect is evaluated on two factors:
    \begin{enumerate}
        \item \textbf{Background scene shape variations:} Similar to object shape variations, the background scene in a generated video also can undergo unnatural morphing or stretching with camera movement.
        \item \textbf{Changing scene with moving camera:} In cases where the camera moves away and returns to the same spot, the entire scene can change, as some generative models only focus on generating the next frames without keeping memory of the scene that was already generated. Even if the background is rigid, unexpected scene change can affect the viewing experience. This dimension is only evaluated if the prompt mentions such a camera motion.
    \end{enumerate}
    \item \textbf{Object-background interaction:} Independent evaluation of foreground objects and background scene can overlook the factors that ground the objects to the actual scene. Thus, we need to evaluate how well the objects interact with the background scene.
    \begin{enumerate}
        \item \textbf{Sliding:} Generative models often find it difficult to keep an object fixed to the ground or platform on which it should be placed, especially when it needs to capture camera motion. Objects tend to slide off the ground, causing humans to prefer them less.
        \item \textbf{Shadow inconsistencies:} The shadows cast by objects must conform to the laws of physics, particularly, should be cast based on the position of the light source and should remain consistent with the shape and motion of the object no matter how the camera moves.
        \item \textbf{Reflection inconsistencies:} Similar to shadow evaluation, reflective surfaces can also be evaluated on how natural the reflections look and how well they reflect nearby objects. Generating reflections under camera motion is more challenging for a generative model, as it needs to regenerate a different view point of a nearby object in good detail.
        
        \textbf{Note:} The dimensions of shadow and reflection consistency are only evaluated when the prompt contains a mention of shadow or reflection.
    \end{enumerate}
    \item \textbf{Camera motion evaluation:} With all the aspects visible in the scene addressed, the last quality aspect that remains to be evaluated is the camera motion itself.
    \begin{enumerate}
        \item \textbf{Unpleasant camera movement:} Videos with unpleasant camera motion can affect the perceptual quality of the video. We ask the subjects to evaluate how unpleasant the perceived camera motion feels.
    \end{enumerate}
\end{enumerate}

Finally we evaluate on an additional dimension that collects a holistic human preference, given the prompt and a pair of videos. In a holistic evaluation, one would select the most preferred video considering various aspects like visual quality, motion quality, overall appeal and text-alignment. Holistic preference data helps in evaluating which individual dimensions contribute more to the final preference. 
\begin{figure*}
  \centering
  \includegraphics[width=0.9\linewidth]{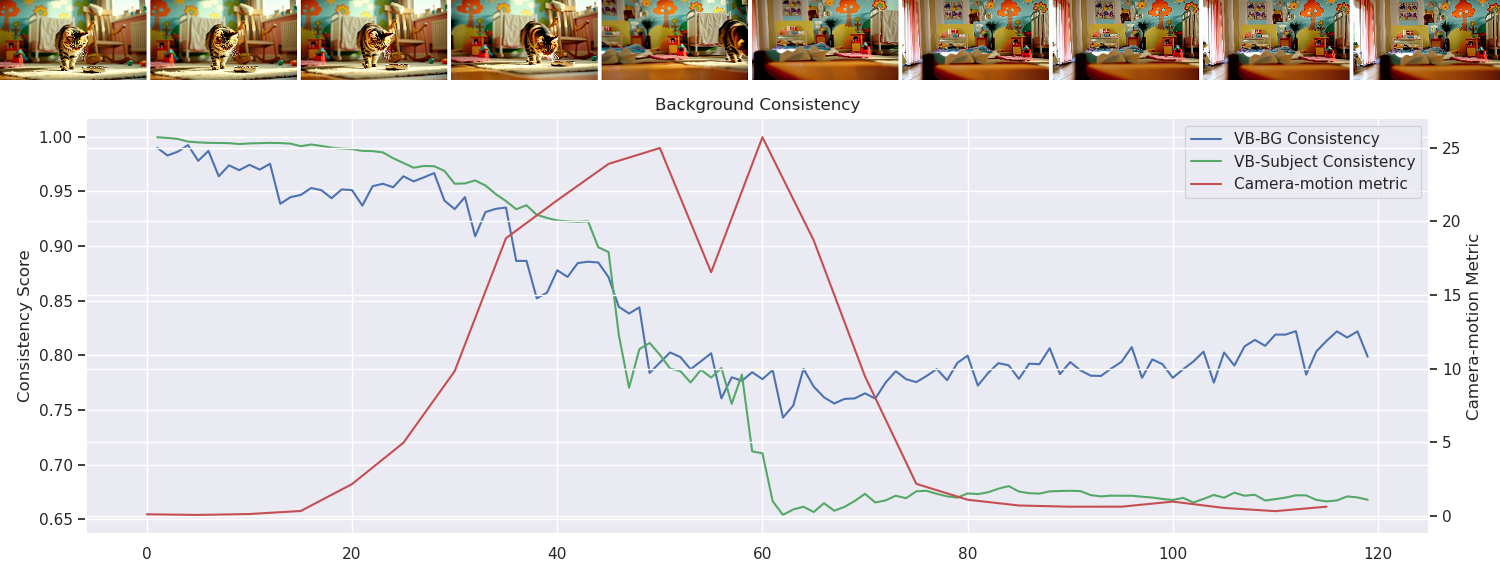}
  \caption{Effect of camera motion on feature-level metrics: We observe that with large camera motions in the scene, the feature level metrics get affected despite having high quality video.}
  \label{fig:feature_static}
\end{figure*}

\section{Dataset Analysis}
\subsection{Prompt Distribution}
We show the percentage of broad categories in each scene element used to construct the prompt suite in Fig.~\ref{fig:dataset_distribution}. In scene setting, there three categories: indoor, outdoor-land, outdoor-water; subjects are broadly classified into human, animal, vehicle, and other objects; for number of subjects, we sample from one, two, three, and many. Additionally, we also show the worcloud generated from our prompt suite in Fig.~\ref{fig:wordcloud}. We observe that a majority of the words focus primarily on camera motion and dynamic scenes.

\subsection{Ratio of dynamic scenes in DynamicEval vs VBench}
\subsubsection{Camera-motion metric}
To identify static and dynamic scenes, we define camera-motion metric using point tracks from CoTracker \cite{cotracker}. We compute point trackers on the entire video using CoTracker and find the variance of each point tracker across frames. All the variances are averaged across points as the final camera-motion metric. Let all the point trackers be denoted as $\{\pointtrack_\pid^f\}_{\pid=1}^\totalpoints$ where $f \in \{1, 2, \cdots, F\}$ represents the frame number. The camera-motion metric is computed as:
\begin{equation}
    C_{\text{cam}} = \frac{1}{P} \sum_{p=1}^{P} \text{Var}\left\{\pointtrack_\pid^f\right\}_{f=1}^F
\end{equation}
Higher $C_{\text{cam}}$, higher the camera motion. 
We do not explicitly mask out foreground objects for this metric. Even a slight camera motion causes all trackers to shift, resulting in a much higher camera motion metric. When the camera is static, most tracks remain stationary even if foreground objects move. Thus, we compute the camera motion metric on all trackers.
We find $\tau_{\text{cam}}$, the threshold that differentiates static vs dynamic by finding the threshold at which the $10\%$ of videos with lowest $C_{\text{cam}}$ are lesser than $\tau_{\text{cam}}$. This selection based on the observation mentioned in Section 3.3 of the main paper. 

\begin{figure}
  \centering
  \includegraphics[width=0.99\linewidth]{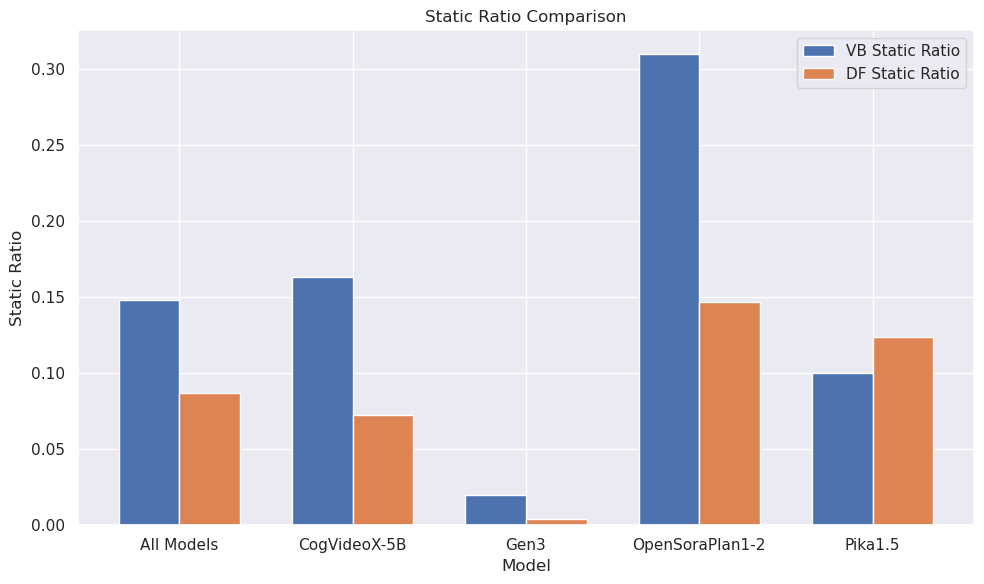}
  \caption{Comparison of ratio static videos in DynamicEval (DE) vs VBench (VB) dataset on matching models. The ratio of static scenes generated from VBench prompts are significantly higher than the ones generated from DynamicEval.}
  \label{fig:vb_static}
\end{figure}

\subsubsection{Comparing with VBench}
We use the camera-motion metric to compare the ratio of static scenes generated from the prompts in VBench with the scenes generated from our DynamicEval. We utilize the videos made available by the VBench authors for CogVideoX, OpenSoraPlan, Gen3 and Pika. We randomly sample unique prompts from the VBench prompt suite and extract the corresponding videos for each model. We compute $C_{\text{cam}}$ on these videos and videos in DynamicEval corresponding to the selected models. We partition each subset into static and dynamic using $\tau_{\text{cam}}$. The ratio of static scenes in both databases across models is shown in Fig.~\ref{fig:vb_static}. The ratio of static scenes in VBench is significantly higher than the ratio of static scenes in DynamicEval, validating the effectiveness of our prompt suite in generating dynamic videos.

\section{Analysis of Baseline Metrics}

\subsection{Bias of feature level metrics with camera motion}
Feature level consistency metrics like VBench background consistency (VB-BG) and subject consistency (VB-SC) are prone to bias from camera motion. To validate this, we analyze frame level plot of the consistency metrics and camera-motion metric in Fig~\ref{fig:feature_static}. The frame level camera-motion metric is calculated by splitting frames into batches of 5 and computing the camera-motion metric for each batch. We select a video with large changes in camera motion and monitor VB-BG, VB-SC, and $C_{\text{cam}}$ at a frame level. We observe that with large camera motion, the feature level metrics tend to be sensitive to such motion in the scene.

\subsection{Failure Cases in baseline metrics}
We provide example video pairs in the supplementary HTML in "Pairwise Comparisons" section. This section shows examples where the baseline metrics fail to capture fine-grained distortions in both background and foreground. In background consistency (Click on "Background Scene Consistency" under "Pairwise Comparisons"), most of the fail cases  of VB-BG correspond to videos with significantly lesser camera motion (also shown in Fig. \ref{fig:fgbg-qual}). Localized distortions are never captured as shown in Example 1 in the page. Even with severe distorions in Video 2 of Example 1, VB-BG still prefers Video 2. All the other examples also reinforce the same point. In foreground object consistency (Click on "Foreground Subject Consistency" under "Pairwise Comparisons"), VB-SC also suffers from high camera motion bias as seen in Examples 1, 3 an d 4. Additionally, there are cases where in case of multiple subjects in a scene VB-SC struggles to evaluate each object for reliable evaluation (Example 2).
\section{Proposed Metrics}
\subsection{Implementation Details}
\subsubsection{MS-Debias: Details of Multi-scale processing}
\begin{figure}

  \centering
  \includegraphics[width=0.6\linewidth]{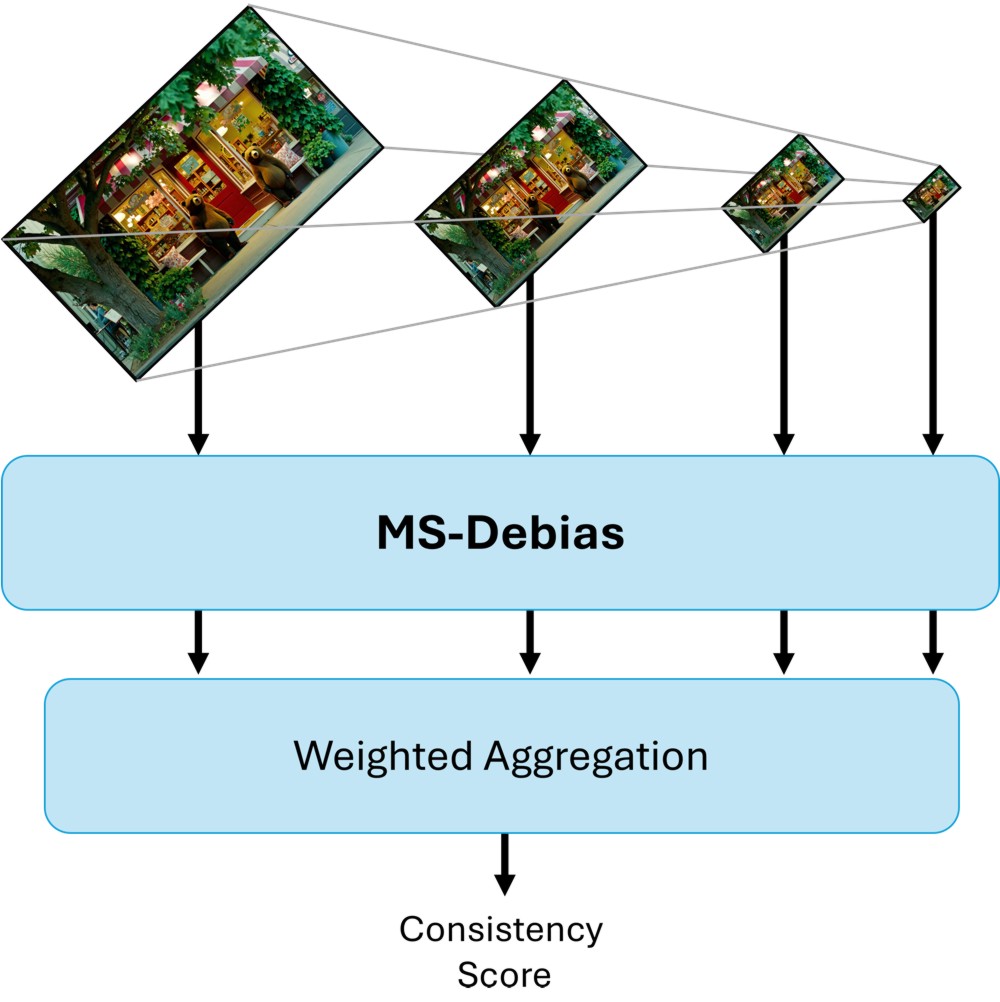}
  \caption{Multi-scale processing in MS-Debias: Different down-sampled versions of the input video are processed independently through the MS-Debias pipeline to obtain background consistency scores at multiple levels. The scores are then averaged using weights proportional to the scales to obtain the final metric score. This mimics the multi-scale processing capabilities of the human visual system.} 
  \label{fig:multi-scale}

\end{figure}
We evaluate MS-Debias metric at multiple scales using Gaussian pyramid down-sampling to create multi-scale videos before feeding them into the interpolation model.
We then apply our debiased motion smoothness pipeline at each scale. To validate the effectiveness of each scale in our MS-Debias metric, we evaluate the performance of our method at different scales as shown in Table \ref{tab:multi-scale}. We report pairwise video preference accuracy for both our method and the baseline motion smoothness metric at each scale. We observe improved performance at lower scales, consistent with prior findings that perceptual video quality is often better characterized at reduced resolutions~\citep{strred}. Finally, incorporating weighted multi-scale processing, our method substantially outperforms the baseline.
We choose the weights for multi-scale processing proportional to the size of the bands. Specifically, we assign weights of $\nicefrac{1}{8}, \nicefrac{1}{4}, \nicefrac{1}{2},$ and $1$ to the original resolution and to videos downscaled by factors of 2, 4, and 8, respectively. We normalize these weights to sum to $1$ before computing the weighted score.
\begin{table}[h]
  \caption{Pairwise video selection accuracy of MS-Debias on DynamicEval (background scene consistency) for each scale in the Gaussian pyramid. We find that, with lower scales the performance increases gradually, and the combination of multiple scales provides the best performance.}
  \label{tab:multi-scale}
  \centering
  \begin{adjustbox}{max width=0.6\textwidth}
  \begin{tabular}{lcc}
    \toprule
    Metrics & VB-MS & MS-Debias\\
    \midrule
    Original resolution & 53.7 & 56.6\\    
    Downscaled by 2 & 55.5 & 57.1\\    
    Downscaled by 4 & 55.7 & 57.3\\    
    Downscaled by 8 & 53.3 & 57.6\\  
    \midrule
    Multi-scale combination & \textbf{56.3} & \textbf{58.2}\\    
    \bottomrule
  \end{tabular}
  \end{adjustbox}
  \end{table}
\subsubsection{Tracker-FG: Additional implementation details}
As we first detect objects using GroundingDINO~\citep{GroundingDINO} and then compute Tracker-FG, there are cases where it does not detect any object in the scene. Therefore, in a given pair, if objects are not detected in both videos, we use the baseline VB-SC metric for comparison. In cases where GroundingDINO detects objects in only one video, our metric selects that particular video as high quality, as videos with no objects are considered bad quality for foreground consistency, as discussed in Section~\ref{sec:FG-GT-Construction}.

\begin{figure*}
  \centering
  \includegraphics[width=0.75\linewidth]{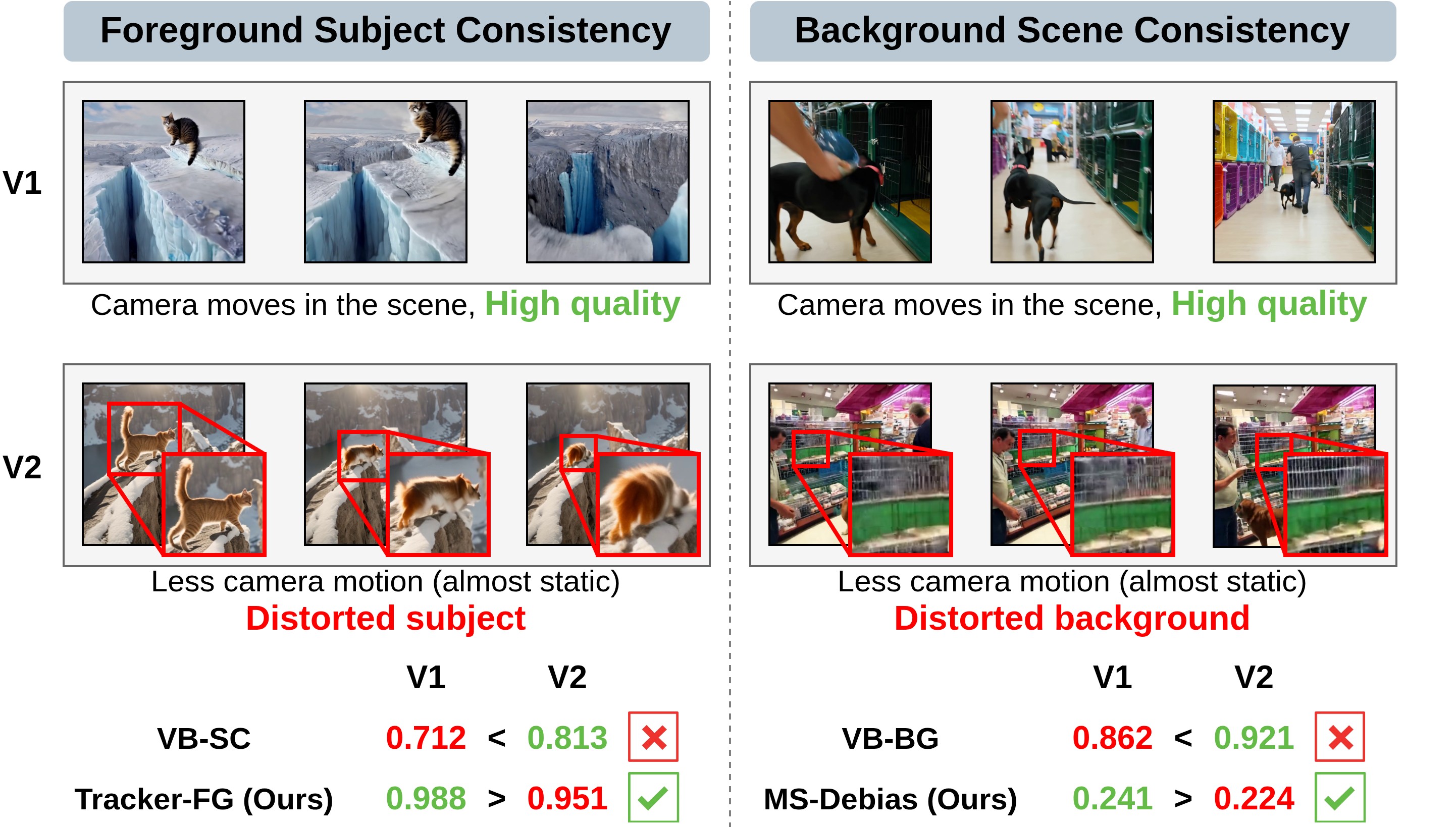}
  \caption{Baseline metrics (VB-SC and VB-BG) exhibit a strong bias with respect to camera motion. As camera motion increases, their scores decrease even for high-quality videos, which is consistent with the visualizations in Fig. \ref{fig:feature_static}. In foreground subject consistency, even though video \textbf{V1} is of high quality and \textbf{V2} shows a cat morphing (distortion), VB-SC gives a lower score to \textbf{V1} because it has a larger camera motion. Similarly, in background consistency, VB-BG gives a low score to \textbf{V1}, which has more camera motion even though \textbf{V2} exhibits significant localized distortions. Our metrics are independent of camera motion bias.}
  \label{fig:fgbg-qual}
\end{figure*}
\vspace{1mm}

\subsubsection{Computational Cost and Practicality in Large Scale Evaluation:}
We have provided a comparison of the average time taken to generate a video using Wan2.1 and compute the evaluation metrics with a single A100 GPU in Table~\ref{tab:timecost}. From the table it is evident that the time taken to evaluate the videos is considerably less than the time taken to generate the video. Additionally, we replace the heavyweight models of SAM2~\citep{sam2024vid} and CoTracker~\citep{cotracker} with their light-weight variants (shown with "-S" as a suffix in the table) to further reduce the time taken for evaluation. Note that in MS-Debias, there is a step that computes video segmentation maps for every objects in the scene to extract clean edges. This mainly contributes to the time cost. With newer lightweight segmentation/tracking models, one can use our framework to further improve the speed. Additionally, there are different parts in the framework that can be parallely computed if optimized for multi-GPU settings.
\begin{table}
  \caption{Time taken on average for generation and evaluation.}
  \label{tab:timecost}
  \centering
  \begin{adjustbox}{max width=0.47\textwidth}
  \begin{tabular}{llll}
  \toprule
    BG Metric & Time & FG Metric & Time \\
    \midrule
    MS-Debias & 15 minutes & Tracker-FG & 2.2 minutes\\    
    MS-Debias-S & 10 minutes & Tracker-FG-S & 1 minute\\    
    \midrule
    Video Generation & 50 minutes &  &  \\  
    \bottomrule
  \end{tabular}
  \end{adjustbox}
\end{table}

\subsection{Analysis}
\subsubsection{Metric Visualizations}
We provide graphical video visualizations of our method to better understand and get an intuition of our metrics in "Visualizing the Metrics" section of our HTML page.

\noindent \textbf{MS-Debias:} We have provided visuals of all the intermediate steps of MS-Debias in the HTML pages. Initially the baseline motion smoothness error map is mostly sensitive to the occlusions/disocclusions ond foreground objects. After adding object and edge maps, the debiased error map tends to show more of the localized issues in the background.

\noindent \textbf{Tracker-FG:} We provide visualization of the average of kNN tracker distance deviation across point tracks per object. We have colored the corresponding object data for convenience. Higher the inconsistency of an object, higher the average deviation. We have also provided a similar visualization for real videos. Here we see that the average deviation stays very low compared to AI videos with object inconsistencies.

\begin{table*}
    \centering
    \caption{Statistical significance tests ($95\%$ confidence level) across all pairs of foreground subject consistency metrics. Table (a) shows the $\chi^2$ and $p$ values of McNemar's statistical test that tests whether the pairwise accuracy differences are significant. A value of $p<0.05$ for a pair of methods means that the accuracy differences are statistically significant. Table (c) shows the low-high bounds of $95\%$ bootstrap percentile confidence interval of accuracy differences. The accuracy differences are statistically significant if the interval does not contain $0$. Table (b) and (d) show the statistical significance matrix, in which, `1': metric in the row has statistically significant performance gains over the metric mentioned in the corresponding column; `0': row metric has statistically significantly poorer performance; `-': the accuracy difference is not statistically significant.}
    \label{tab:stat-sig-FG}

    \begin{subtable}[t]{0.27\textwidth}
        \centering
        \caption{$\chi^2$ ($p$) \\values}
        \label{tab:FG-mcnemar-chi-p}
        \begin{adjustbox}{max width=\textwidth}
        \begin{tabular}{lcc}
        \toprule
        &VB-SC&Tracker-FG\\
        \midrule
        VB-SC&\cellcolor{gray!20} & 27.5 (1.6e-07)\\
        Tracker-FG&27.5 (1.6e-07) & \cellcolor{gray!20} \\
        \bottomrule
        \end{tabular}
        \end{adjustbox}
    \end{subtable}
    \hfill
    \begin{subtable}[t]{0.22\textwidth}
        \centering
        \caption{McNemar's \\statistical significance}
        \label{tab:FG-mcnemar-significance}
        \begin{adjustbox}{max width=\textwidth}
        \begin{tabular}{lcc}
        \toprule
        &VB-SC&Tracker-FG\\
        \midrule
        VB-SC&\cellcolor{gray!20} & 0 \\
        Tracker-FG&1 & \cellcolor{gray!20} \\
        \bottomrule
        \end{tabular}
        \end{adjustbox}
    \end{subtable}
    \hfill
    \begin{subtable}[t]{0.27\textwidth}
        \centering
        \caption{Bootstrap CI \\low-high intervals}
        \label{tab:FG-bCI}
        \begin{adjustbox}{max width=\textwidth}
        \begin{tabular}{lcc}
        \toprule
        &VB-SC&Tracker-FG\\
        \midrule
        VB-SC&\cellcolor{gray!20} & (-0.027, -0.012)\\
        Tracker-FG&(0.012, 0.027) & \cellcolor{gray!20} \\
        \bottomrule
        \end{tabular}
        \end{adjustbox}
    \end{subtable}
    \hfill
    \begin{subtable}[t]{0.22\textwidth}
        \centering
        \caption{Bootstrap CI \\statistical significance}
        \label{tab:FG-bCI-significance}
        \begin{adjustbox}{max width=\textwidth}
        \begin{tabular}{lcc}
        \toprule
        &VB-SC&Tracker-FG\\
        \midrule
        VB-SC&\cellcolor{gray!20} & 0 \\
        Tracker-FG&1 & \cellcolor{gray!20} \\
        \bottomrule
        \end{tabular}
        \end{adjustbox}
    \end{subtable}

\end{table*}

\begin{table*}
    \centering
    \caption{Statistical significance tests ($95\%$ confidence level) across all pairs of background consistency metrics. Details are already discussed in Table \ref{tab:stat-sig-FG}.
    \label{tab:stat-sig-BG}}

    \begin{subtable}[t]{0.56\textwidth}
        \centering
        \caption{$\chi^2$ ($p$) \\values}
        \label{tab:BG-mcnemar-chi-p}
        \begin{adjustbox}{max width=\textwidth}
        \begin{tabular}{lcccccc}
        \toprule
        &VB-Temp.Flick.&EC-Sem&VB-MS&MEt3R&VB-BG&MS-Debias\\	
        \midrule
        VB-Temp.Flick.&\cellcolor{gray!20} & 1.1 (2.9e-01) & 17.7 (2.6e-05) & 40.3 (2.2e-10) & 43.6 (4.1e-11) & 102.5 (4.2e-24)\\
        EC-Sem&1.1 (2.9e-01) & \cellcolor{gray!20} & 6.5 (1.1e-02) & 21.4 (3.7e-06) & 31.9 (1.6e-08) & 78.5 (8.2e-19)\\
        VB-MS.& 17.7 (2.6e-05) & 6.5 (1.1e-02) & \cellcolor{gray!20} & 3.6 (5.8e-02) & 10.1 (1.5e-03) & 52.2 (5.0e-13)\\
        MEt3R&40.3 (2.2e-10) & 21.4 (3.7e-06) & 3.6 (5.8e-02) & \cellcolor{gray!20} & 2.5 (1.1e-01) & 25.5 (4.4e-07)\\
        VB-BG&43.6 (4.1e-11) & 31.9 (1.6e-08) & 10.1 (1.5e-03) & 2.5 (1.1e-01) & \cellcolor{gray!20} & 26.2 (3.1e-07)\\
        MS-Debias&102.5 (4.2e-24) & 78.5 (8.2e-19) & 52.2 (5.0e-13) & 25.5 (4.4e-07) & 26.2 (3.1e-07) & \cellcolor{gray!20} \\
        \bottomrule
        \end{tabular}
        \end{adjustbox}
    \end{subtable}
    \hfill
    \begin{subtable}[t]{0.43\textwidth}
        \centering
        \caption{McNemar's \\statistical significance}
        \label{tab:BG-mcnemar-significance}
        \begin{adjustbox}{max width=\textwidth}
        \begin{tabular}{lcccccc}
        \toprule
        &VB-Temp.Flick.&EC-Sem&VB-MS&MEt3R&VB-BG&MS-Debias\\	
        \midrule
        VB-Temp.Flick.&\cellcolor{gray!20} & - & 0 & 0 & 0 & 0\\
        EC-Sem& - & \cellcolor{gray!20} & 0 & 0 & 0 & 0\\
        VB-MS.& 1 & 1 & \cellcolor{gray!20} & - & 0 & 0\\
        MEt3R&1 & 1 & - & \cellcolor{gray!20} & - & 0\\
        VB-BG&1 & 1 & 1 & - & \cellcolor{gray!20} & 0\\
        MS-Debias&1 & 1 & 1 & 1 & 1 & \cellcolor{gray!20}\\
        \bottomrule
        \end{tabular}
        \end{adjustbox}
    \end{subtable}
    \hfill
    \begin{subtable}[t]{0.56\textwidth}
        \centering
        \caption{Bootstrap CI \\low-high intervals}
        \label{tab:BG-bCI}
        \begin{adjustbox}{max width=\textwidth}
        \begin{tabular}{lcccccc}
        \toprule
        &VB-Temp.Flick.&EC-Sem&VB-MS&MEt3R&VB-BG&MS-Debias\\	
        \midrule
        VB-Temp.Flick.&\cellcolor{gray!20} & (-0.018, 0.005) & (-0.034, -0.013) & (-0.046, -0.024) & (-0.060, -0.032) & (-0.082, -0.056)\\
        EC-Sem& (-0.005, 0.018) & \cellcolor{gray!20} & (-0.030, -0.004) & (-0.041, -0.016) & (-0.054, -0.026) & (-0.076, -0.049)\\
        VB-MS.& (0.013, 0.034) & (0.004, 0.030) & \cellcolor{gray!20} & (-0.024, 0.000) & (-0.037, -0.009) & (-0.058, -0.033)\\
        MEt3R&(0.024, 0.046) & (0.016, 0.041) & (-0.000, 0.024) & \cellcolor{gray!20} & (-0.024, 0.003) & (-0.047, -0.021)\\
        VB-BG&(0.032, 0.060) & (0.026, 0.054) & (0.009, 0.037) & (-0.003, 0.024) & \cellcolor{gray!20} & (-0.031, -0.014)\\
        MS-Debias&(0.056, 0.082) & (0.049, 0.076) & (0.033, 0.058) & (0.021, 0.047) & (0.014, 0.031) & \cellcolor{gray!20}\\
        \bottomrule
        \end{tabular}
        \end{adjustbox}
    \end{subtable}
    \hfill
    \begin{subtable}[t]{0.43\textwidth}
        \centering
        \caption{Bootstrap CI \\statistical significance}
        \label{tab:BG-bCI-significance}
        \begin{adjustbox}{max width=\textwidth}
        \begin{tabular}{lcccccc}
        \toprule
        &VB-Temp.Flick.&EC-Sem&VB-MS&MEt3R&VB-BG&MS-Debias\\	
        \midrule
        VB-Temp.Flick.&\cellcolor{gray!20} & - & 0 & 0 & 0 & 0\\
        EC-Sem& - & \cellcolor{gray!20} & 0 & 0 & 0 & 0\\
        VB-MS.& 1 & 1 & \cellcolor{gray!20} & - & 0 & 0\\
        MEt3R&1 & 1 & - & \cellcolor{gray!20} & - & 0\\
        VB-BG&1 & 1 & 1 & - & \cellcolor{gray!20} & 0\\
        MS-Debias&1 & 1 & 1 & 1 & 1 & \cellcolor{gray!20}\\
        \bottomrule
        \end{tabular}
        \end{adjustbox}
    \end{subtable}

\end{table*}

\subsubsection{Qualitative Analysis}
\label{sec:qual}

We qualitatively analyze our metrics in "Pairwise Comparisons" section in HTML. For background consistency, VBench background consistency (VB-BG) fails to predict the higher quality videos especially when there is lesser camera motion. VB-BG tends to reject videos with more camera motion. Whereas MS-Debias is able to prefer the correct example in such cases by focusing on the localized distortions (Fig. \ref{fig:fgbg-qual}) (Refer to all examples in "Pairwise Comparisons" - "Background Scene Consistency" in HTML). The same issue is seen in VB-Subject consistency (VB-SC), with VB-SC prefering videos with lower motion more (Fig. \ref{fig:fgbg-qual}). Tracker-FG focuses on the object without getting biased by camera or object motion. It captures long term dependencies more effectively and is capable of tracking each subject individually to computes consistency, which is missing in VB-SC.

We further analyze some failure cases where both baseline and proposed metrics fail to select the higher quality video. MS-Debias tend fail when the generated video has very graduall variations in background. The flow based method is able to reconstruct  the intermediate frames even though the videos look unnatural. Humans tend to prefer more natural scenes. A similar trend is seen with Tracker-FG. Humans reject unnatural objects even if they look rigid and flow smoothly. 




\subsubsection{Statistical Significance Tests}
\label{sec:statsig}

\noindent \textbf{McNemar's Test:} To evaluate the significance of differences between metrics on the same test instances, we apply McNemar's test~\cite{McNemar} across all pairs of metrics $(i, j)$ for both evaluation dimensions. For each pair, we form the standard $2\times2$ contingency table of agreement and disagreement counts, and compute the McNemar chi-square statistic $\chi^2_{i,j}$ with Yates’ continuity correction and its corresponding p-value $p_{i,j}$. McNemar’s test evaluates the null hypothesis that the two methods have equal proportions of correct predictions (i.e., equal accuracy) and that the observed disagreements (A correct vs. B correct) are symmetric. A small p-value indicates that the observed asymmetry is unlikely to have occurred by chance. We consider two metrics to be significantly different at the $95\%$ confidence level (i.e., $p_{i,j} < 0.05$). We report the chi-square statistics for all pairs of metrics for foreground consistency along with their p-values in Table \ref{tab:FG-mcnemar-chi-p} and the statistical significance matrix in Table \ref{tab:FG-mcnemar-significance}. Similarly, we report the same for background consistency in Table \ref{tab:BG-mcnemar-chi-p} and Table \ref{tab:BG-mcnemar-significance}. We order the metrics such that their pairwise preference performance increases monotonically for better readability. A value of `1' in the significance matrix indicates that the metric in the row is significantly better than the metric in the corresponding column, `0' means that it is significantly worse and `-' means that the difference in metrics is statistically insignificant. The results show that our metrics (Tracker-FG and MS-Debias) are statistically superior to all other metrics across both evaluation dimensions, with consistently low p-values. Additionally, MEt3R is statistically similar to VB-Motion Smoothness, which is consistent with the fact that both metrics attempt to predict intermediate frames from reconstructions. VB-BG is also statistically similar to MEt3R, and VB-Temporal Flickering is similarly close to EC-Semantic Consistency.

\noindent \textbf{Bootstrap Confidence Intervals:} To quantify the uncertainty in accuracy differences between methods, we compute paired nonparametric bootstrap confidence intervals (CIs)~\cite{bootstrapCI} with $95\%$ confidence level ($T=10000$ resamples). For each pair of methods $(i,j)$, we report the bootstrap low–high interval for the accuracy difference
\begin{equation}
    \Delta_{i,j} = \text{Acc}(i) - \text{Acc}(j)
\end{equation}
is shown directly in Table \ref{tab:FG-bCI} and \ref{tab:BG-bCI}, where each cell contains the [low, high] interval obtained from the bootstrap distribution. A pairwise difference is considered statistically significant at the $95\%$ confidence level if the corresponding interval does not include 0. Based on this criterion, Table \ref{tab:FG-bCI-significance} and Table \ref{tab:BG-bCI-significance} summarizes the significance matrices for every pair of methods. Consistent with McNemar's significance tests, the Bootstrap CIs also indicate that our proposed metrics are statistically superior to all other metrics.

Bootstrap CIs complement McNemar’s test by quantifying the magnitude and uncertainty of the accuracy difference itself rather than just checking whether the direction of pairwise disagreements is statistically significant.

\subsubsection{Reliability of Model-Level Correlations}
Because the model-level correlations are computed over only $n{=}10$
generators, we characterize their statistical reliability with two complementary analyses Table~\ref{tab:model_level_reliability}. 
First, we report $95\%$ confidence
intervals from a case-resampling bootstrap ($B{=}10{,}000$ resamples of the ten models with
replacement, percentile method). Second, as a direct outlier check, we perform a leave-one-model-out
(LOO) analysis, recomputing PLCC with each generator removed in turn and reporting the range. Under
both, Tracker-FG and MS-Debias remain strong: their PLCC confidence intervals exclude zero
($p{=}0.009$ and $p{=}0.014$) and their LOO ranges stay high and narrow ($[0.74,0.85]$ and
$[0.70,0.83]$), indicating that no single generator drives the correlation. The VBench baselines are
weaker and less stable: VB-SC is not significant (PLCC $0.33$, $p{=}0.35$, CI spanning zero) and VB-BG
is borderline ($p{=}0.10$) with its LOO range dropping to $0.49$. We emphasize that at $n{=}10$ the
bootstrap intervals are necessarily wide and partially overlap across methods; the LOO analysis and
the consistently higher point estimates (rather than non-overlapping CIs) are the primary evidence
that our metrics track human model preferences more reliably.

\begin{table}[t]
\centering\small
\caption{Reliability of model-level win-ratio correlations ($n=10$ generators, full-agreement
subset). 95\% CIs from a $B=10{,}000$ case-resampling bootstrap; LOO-PLCC is the range of PLCC under
leave-one-model-out jackknife.}
\label{tab:model_level_reliability}
\begin{adjustbox}{max width=0.47\textwidth}
\begin{tabular}{llcc}
\toprule
Dim. & Method & PLCC [95\% CI] ($p$) & LOO-PLCC \\
\midrule
\multirow{2}{*}{FG}
 & VB-SC       & 0.334 [-0.31, 0.83] (.345) & [0.17, 0.53] \\
 & \textbf{Tracker-FG} & \textbf{0.772 [0.46, 0.94] (.009)} & \textbf{[0.74, 0.85]} \\
\midrule
\multirow{2}{*}{BG}
 & VB-BG       & 0.551 [0.15, 0.92] (.099) & [0.49, 0.71] \\
 & \textbf{MS-Debias} & \textbf{0.743 [0.44, 0.96] (.014)} & \textbf{[0.70, 0.83]} \\
\bottomrule
\end{tabular}
\end{adjustbox}
\end{table}

\subsubsection{Isolated Metric Ablation}

Our final foreground and background metrics (Tracker-FG and MS-Debias) each fuse a pixel-level consistency signal with the corresponding VBench consistency score (subject and background, respectively). To disentangle the contributions of each component, we report each metric in isolation before fusion. As shown in Table~\ref{tab:isolated_ablation}, the isolated signals already outperform their VBench counterparts on both the full data and the full-agreement subset. Fusing the two signals yields a further, larger improvement, showing that the pixel-level signal and the VBench consistency score are complementary rather than redundant, and that neither component alone accounts for the performance of the final metric.

\begin{table}[t]
\centering
\caption{Pairwise-agreement accuracy of each metric in isolation (w/o fusion) versus the final fused metric, reported on the full data and the full-agreement annotation subset. The pixel-level signal alone already beats the VBench baseline, and fusion yields a further gain.}
\label{tab:isolated_ablation}
\begin{adjustbox}{max width=0.47\textwidth}
\begin{tabular}{lcc}
\toprule
Method & Full Dataset & Full Agreement \\
\midrule
\multicolumn{3}{c}{Subject Consistency (Pairwise Acc.)} \\
\midrule
VB-SC                        & 56.2 & 58.8 \\
Tracker-FG w/o fusion        & 57.1 & 59.8 \\
\textbf{Tracker-FG (ours)}   & \textbf{58.2} & \textbf{62.7} \\
\midrule
\multicolumn{3}{c}{Background Consistency (Pairwise Acc.)} \\
\midrule
VB-BG                        & 56.0 & 59.3 \\
VB-MS                        & 53.7 & 56.8 \\
MS-Debias w/o fusion         & 56.3 & 59.6 \\
\textbf{MS-Debias (ours)}    & \textbf{58.2} & \textbf{62.7} \\
\bottomrule
\end{tabular}
\end{adjustbox}
\end{table}

\subsubsection{Pairwise preference on inter-model vs intra-model comparisons}
We evaluate how well our metrics distinguish between videos generated by different models and those generated by the same model by comparing the pairwise preference accuracy of the metrics on the inter- and intra-model subsets in Table \ref{tab:inter-intra}. In both inter-model and intra-model comparisons, our proposed metrics outperform the baselines. As intra-model comparisons involve comparing videos with similar content and resolution, the accuracy is generally higher.

\begin{table}
  \caption{Performance of the methods on inter-model vs intra-model video comparisons. The proposed metrics perform better on both comparisons, with higher performance on intra-model pairs, as the videos contain similar content.}
    \centering
    


    \begin{tabular}{lcc}
        \toprule
        & \multicolumn{2}{c}{Background Consistency}\\
        \midrule
        Comparison Type & VB-BG & MS-Debias\\
        \midrule
        Inter-model Comparisons & 54.8 & \textbf{58.3}\\
        Intra-model Comparisons & 57.7 & \textbf{58.2}\\
        \midrule
        & \multicolumn{2}{c}{Subject Consistency}\\
        \midrule
        & VB-SC & Tracker-FG\\
        \midrule
        Inter-model Comparisons & 54.3 & \textbf{56.5} \\
        Intra-model Comparisons & 59.1 & \textbf{60.7} \\
        \bottomrule
    \end{tabular}
    
  \label{tab:inter-intra}
\end{table}

\subsubsection{Comparison with models fine-tuned on subjective quality}
VBench and EvalCrafter provides more metrics that use models fine-tuned using quality labels on real videos, such as VB-quality~\citep{vbench} and EC-Dover~\citep{evalcrafter}. VB-quality uses the MUSIQ~\citep{ke2021musiq} image quality predictor which is trained on camera captured images with subjective quality annotations. EC-Dover uses the DOVER~\citep{wu2023exploring} video quality prediction model, which is trained on real-world distorted videos, where human subjects annotate them on perceptual quality. In contrast, our metrics are purely zero-shot with respect to quality evaluation. We compare our background consistency metric with quality pre-trained metrics proposed in VBench and EvalCrafter in Table~\ref{tab:qa-models}. The fine-tuned metrics generally outperform the baseline metric VB-BG, with EC-Dover achieving the best results, likely due to being trained on videos. Notably, despite being a zero shot metric, MS-Debias achieves the same performance best metric fine-tuned on subjective quality labels.

\begin{table}
  \caption{Pairwise preference accuracy compared with models trained on subjective quality.}
  \label{tab:qa-models}
  \centering
  \begin{adjustbox}{max width=0.47\textwidth}
  \begin{tabular}{l|cc|cc}
  \toprule
    Method & VB-Quality & EC-Dover & VB-BG & MS-Debias \\
    \midrule
    Accuracy & 56.5 & \textbf{58.2} & 56.0 & \textbf{58.2}\\         
    \bottomrule
  \end{tabular}
  \end{adjustbox}
\end{table}

\subsubsection{Mutual overlap between both metrics}
In the background scene consistency metric MSDebias, we provide an object mask to debias the effect of foreground objects in the computation. Similarly, for the foreground consistency metric Tracker-FG, we explicitly compute the tracks only in the object mask regions. Therefore, in both the metrics there is clearly no overlap in terms of the features they evaluate. In contrast, the baseline VBench metrics do not explicitly separate out the foreground and background regions as they are computed on holistic deep features. Neverthless, to validate that there is no overlap in performance, we evaluate the pairwise preference accuracy of MS-Debias on foreground consistency and Tracker-FG on background consistency, swapping the metrics, as shown in the Table \ref{tab:swap-metric}. The poor performance of each measure in the last row indicates how well separated the metrics are.

\begin{table}
    \centering
    \caption{Pairwise preference accuracy after swapping the metrics}
    \label{tab:swap-metric}
    \begin{adjustbox}{max width=0.4\textwidth}
    \begin{tabular}{lcc}
    \toprule
    Method & Full Dataset & Full Agreement \\
    \midrule
    \multicolumn{3}{c}{Background Consistency (Pairwise Acc.)} \\
    \midrule
    VB-BG & 56.0 & 59.3 \\
    MS-Debias & \textbf{58.2} & \textbf{62.7} \\
    \midrule
    Tracker-FG$^*$ & 52.7 & 54.4 \\
    \midrule
    \multicolumn{3}{c}{Subject Consistency (Pairwise Acc.)} \\
    \midrule
    VB-SC & 56.2 & 58.8 \\
    Tracker-FG & \textbf{58.2} & \textbf{62.7} \\
    \midrule
    MS-Debias$^*$ & 53.7 & 53.7 \\
    \bottomrule
    \end{tabular}
    \end{adjustbox}
\end{table}
\section{Miscellaneous}
\subsection{Evaluation Metrics}
\textbf{Top-k video selection accuracy:} Top-k video selection accuracy is the proportion of times the highest quality video is among the top-k predicted videos from the metric. For Top-k evaluation, we obtain ground truth ranking of videos given a prompt through win ratios.
To obtain the ground truth ranking of videos in a prompt, we first filter out the pairwise comparisons between all combinations of $10$ videos each selected from each model to obtain $^{10}C_2=45$ pairs. For each video, there will be $10-1=9$ pair comparisons from which we compute the win ratio of the video as the number of times the video gets selected among $9$ pairs. The video with the highest win ratio is considered as the highest preferred video.

Note that we do not report Spearman's rank order correlation, Pearson's linear correlation or Kendall's rank correlation at a video level, as we have collected a prompt conditioned pairwise video comparison database and not a database with absolute score per video. Rank correlations at a video level can only be applied to an overall video level score that is not conditioned on any variable (eg. prompt).

\section{Broader Impact and Limitations}
DynamicEval enables research on fine-grained evaluation of text-to-video models ensuring human preference alignment on a video level. This promotes development of metrics that correlates well with human preference not only to select better T2V models, but also to select better videos given a pool of candidate videos from the best T2V models, leading to high-quality video content in real-world applications. Our interpretable design of metrics enables researchers to localize and mitigate quality issues in generated videos. Better evaluation tools can lead to higher quality content generation, which can be misused for malicious purposes (generating realistic fake videos). This raises the need for responsible deployment and regulations.

In addition to risks from generative model misuse, we identify two potential scenarios where our evaluation metrics could themselves be misused. First, if our metrics are used for real/fake detection, they may incorrectly classify high-quality generated videos as real. We therefore recommend using these metrics in conjunction with established real/fake detectors. Another potential for misuse is in evaluating and selecting generative models based on these metrics, that can be prone to over-optimization. Therefore, it is always ideal to mix multiple metrics (and potentially keep adding newer metrics) for a more holistic evaluation of the generated videos.


\end{document}